\documentclass{article}
\usepackage{ijcai23}

\usepackage{times}
\usepackage{soul}
\usepackage{url}
\usepackage[hidelinks]{hyperref}
\usepackage[utf8]{inputenc}
\usepackage[small]{caption}
\usepackage{graphicx}
\usepackage{amsmath}
\usepackage{amsthm}
\usepackage{amssymb}
\usepackage{nicefrac}
\usepackage{float}
\usepackage{dblfloatfix}
\usepackage{booktabs}
\usepackage[switch]{lineno}

\usepackage{algpseudocode}
\usepackage[ruled, lined, linesnumbered, commentsnumbered, longend]{algorithm2e}

\usepackage{tikz}
\usetikzlibrary{arrows,arrows.meta,backgrounds,decorations.pathmorphing,positioning,fit,trees,shapes,shadows,automata,calc}
\usetikzlibrary{shapes.geometric}
\usepackage{subfig}
\usepackage{graphicx}

\usepackage{relsize}

\usepackage{mdframed}

\usepackage{comment}
\usepackage{pgfplots}

\newtheorem{example}{Example}

\newtheorem{definition}{Definition}
\newtheorem{proposition}{Proposition}
\definecolor{color1e}{RGB}{55,126,184} 
\definecolor{color2e}{RGB}{228,26,28} 
\definecolor{color3e}{RGB}{77,175,74} 
\definecolor{color1}{RGB}{105,105,105} 
\definecolor{color2}{RGB}{128,128,0}
\definecolor{color3}{RGB}{9, 121, 105}  
\definecolor{color4}{RGB}{139,0,139} 
\definecolor{color5}{RGB}{255,127,0} 
\definecolor{color6}{RGB}{0,0,117} 
\definecolor{color8}{RGB}{0.2, 0.2, 0.2} 
\definecolor{color7}{RGB}{0.48, 0.25, 0.0} 
\definecolor{color9}{RGB}{0,100,0}
\definecolor{commentcolor}{RGB}{60,114,26}
\definecolor{blue-pigment}{rgb}{0.2, 0.2, 0.6}
\definecolor{dark-orange}{RGB}{230,116,81}

\usepackage{cleveref}

\newlength{\scatterplotsize}
\newcommand\dboxed[1]{\tikz [baseline=(boxed word.base)] \node (boxed word) [draw, rectangle, dashed, line cap=round,color=blue-pigment] {#1};}

\newcommand{\covidyes}{0.05}
\newcommand{\covidno}{0.95}

\newcommand{\sympyescovidyes}{0.698}
\newcommand{\sympnocovidyes}{0.302}

\newcommand{\sympyescovidno}{0.1}
\newcommand{\sympnocovidno}{0.9}

\newcommand{\antigenposYY}{0.72}
\newcommand{\antigennegYY}{0.28}

\newcommand{\antigenposYYPAR}{p}
\newcommand{\antigennegYYPAR}{1-p}

\newcommand{\antigenposYN}{0.58}
\newcommand{\antigennegYN}{0.42}
\newcommand{\antigenposNY}{0.005}
\newcommand{\antigennegNY}{0.995}
\newcommand{\antigenposNN}{0.01}
\newcommand{\antigennegNN}{0.99}

\newcommand{\PCRposCOVIDYPAR}{q}
\newcommand{\PCRnegCOVIDYPAR}{1-q}

\newcommand{\PCRposCOVIDY}{0.95}
\newcommand{\PCRnegCOVIDY}{0.05}
\newcommand{\PCRposCOVIDN}{0.04}
\newcommand{\PCRnegCOVIDN}{0.96}

\DeclareMathOperator*{\argmin}{\arg\!\min}

    \SetKwFunction{PSP}{PLA}
	\SetKwFunction{FminimalInst}{getMinDistInst}
    \SetKwFunction{NotFeas}{Infeasible}
   \SetKwFunction{NotFound}{Not found}

 	\SetKwFunction{ArgMin}{argmin}
 	 \SetKwFunction{EC}{EC}

    \SetKwInOut{KwIn}{Input}
    \SetKwInOut{KwOut}{Output}
    
    \SetKwFunction{FMain}{Main}
    \SetKwFunction{FEpPartition}{$\epsilon$-partitioning}
        \SetKwFunction{tuning}{minChgTuning}
        \SetKwFunction{computeMC}{computeMC}
        \SetKwFunction{reachSpec}{reachSpec}
     \SetKwFunction{expandRegionCD}{makeRegion-CD}
     \SetKwFunction{expandRegionEC}{makeRegion-EC}
    \SetKwFunction{expandRegionD}{makeRegion-d}

  \SetKwProg{Fn}{function}{:}{}

\mathchardef\mhyphen="2D

\newcommand{\B}{\mathcal{B}}
\newcommand{\M}{\mathcal{M}}

\newcommand{\Distr}{\mbox{\it Distr}}
\newcommand{\pDistr}{\mbox{\it pDistr}}

\newcommand{\found}{\mbox{\it \scriptsize found}}
\newcommand{\Eval}{\mbox{ \footnotesize Eval}}
\newcommand{\CD}{\mbox{\it CD}}

\newcommand{\modif}{\mbox{\it \scriptsize modif}}

\newcommand{\storm}{\mbox{\it \scriptsize Storm}}
\newcommand{\bayesserver}{\mbox{\it \scriptsize Bayes}}
\newcommand{\samiam}{\mbox{\it \scriptsize SamIam}}

\newcommand{\antigenposNNPAR}{$\textcolor{black}{t}$}
\newcommand{\antigennegNNPAR}{$\textcolor{black}{1{-}t}$}

\newcommand{\antigenposNNMin}{0.0075}
\newcommand{\antigenposNNMax}{0.0125}

\newcommand{\antigennegNNMin}{0.9925}
\newcommand{\antigennegNNMax}{0.9875}

\newcommand{\PCRposCOVIDNMin}{0.0075}
\newcommand{\PCRposCOVIDNMax}{0.0125}

\newcommand{\PCRnegCOVIDNMin}{0.9925}
\newcommand{\PCRnegCOVIDNMax}{0.9875}

\title{Finding an $\epsilon$-Close Minimal Variation of
Parameters in Bayesian Networks}

\author{
Bahare Salmani  and Joost-Pieter Katoen  
	\affiliations
    RWTH Aachen University
    \emails
    salmani,katoen@cs.rwth-aachen.de
}

\begin{document}

\maketitle

\begin{abstract}
  This paper addresses the $\epsilon$-close parameter tuning problem for Bayesian
networks (BNs): find a minimal $\epsilon$-close amendment of probability entries
in a given set of (rows in) conditional probability tables that make a
given quantitative constraint on the BN valid. Based on the
state-of-the-art “region verification” techniques for parametric Markov
chains, we propose an algorithm whose capabilities go
beyond any existing techniques. Our experiments show that $\epsilon$-close tuning
of large BN benchmarks with up to eight parameters is feasible. In
particular, by allowing (i) varied parameters in multiple CPTs and (ii)
inter-CPT parameter dependencies, we treat subclasses of parametric BNs
that have received scant attention so far.
\end{abstract}

\section{Introduction}
 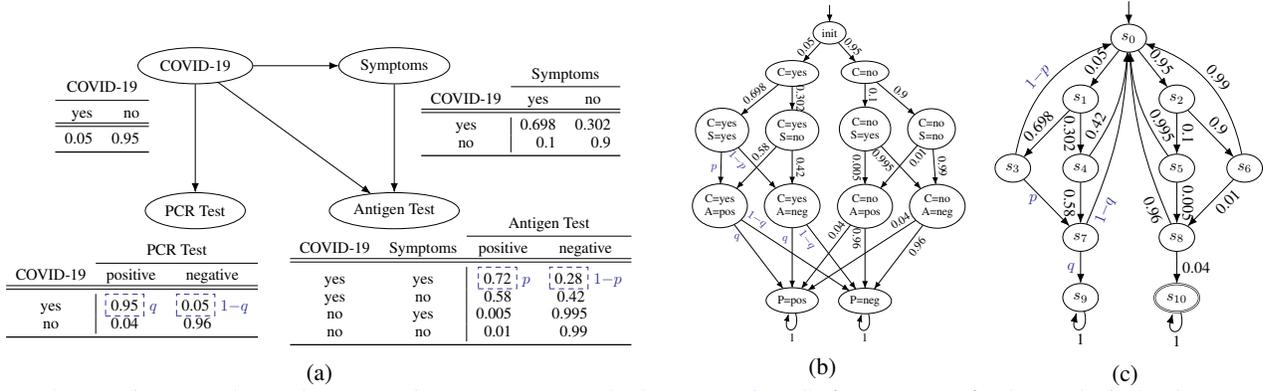
\begin{figure*}[btp]
     \vspace*{-0.8cm}
     \centering
       \begin{minipage}{0.05\linewidth}
  \end{minipage}
  \hfill
   \subfloat[]{
   	\label{fig:pBN-example}
   \centering
     \begin{minipage}{0.5\linewidth}
     \centering
      \resizebox{0.6\width }{0.6\height}{%
       	\begin{tikzpicture}
	[
	node distance=1cm and 0.5cm,
	mNode/.style={draw,ellipse,align=center, minimum size=0.8cm}
	]
	\node[state,draw=none] (help){};
	\node[mNode,below right=0.6cm and 1cm of help] (v0) {Symptoms};
	\node[mNode,below left=0.6cm and 1cm of help] (v1) {COVID-19};
	\node[mNode,below=2.4cm of v1] (v2) {PCR Test};
	\node[mNode,below=2.4cm of v0] (v3) {Antigen Test};
	\node[below right=-0.4cm and -0.4cm of v0] (tabSymp)
	{
		\begin{tabular}{c|rr}
        \multicolumn{1}{c}{} & \multicolumn{2}{c}{Symptoms} \\ 
        \cmidrule{2-3}
        \multicolumn{1}{c}{COVID-19} & \multicolumn{1}{c}{yes} & \multicolumn{1}{c}{no} \\ 
        \hline\midrule
        yes & \sympyescovidyes & \sympnocovidyes \\
        no & \sympyescovidno & \sympnocovidno\\
        \bottomrule
        \end{tabular}
	};
	\node[below left=-0.1cm and 0cm of v1] (tabCOVID)
	{
		\begin{tabular}{rr}
        \multicolumn{2}{c}{COVID-19} \\ 
        \midrule
        yes & no \\ 
        \hline\midrule
        \multicolumn{1}{r}{\covidyes} & \multicolumn{1}{r}{\covidno} \\
        \bottomrule
        \end{tabular}
	};
	
	\node[below left=-0.3cm and -6.4cm of v3] (tabAntigen)
	{
		\begin{tabular}{cc|rr}
            & \multicolumn{1}{c}{} & \multicolumn{2}{c}{Antigen Test}  \\ 
            \cmidrule{3-4}
            COVID-19 & \multicolumn{1}{c}{Symptoms} & \multicolumn{1}{c}{positive} & \multicolumn{1}{c}{negative}          \\ 
            \hline\midrule
            yes  & yes    & \dboxed{\textcolor{black}{\antigenposYY}}$\,\,\textcolor{blue-pigment}{p}$ & \dboxed{\textcolor{black}{\antigennegYY}}$\,\,\textcolor{blue-pigment}{1{-}p}$  \\
            yes & no    & \antigenposYN\,\,\,\,\,\,\,
             & \antigennegYN\,\,\quad\quad\quad     \\
            no  & yes   & \antigenposNY\,\,\,\,\,\,\, & \antigennegNY\,\quad\quad\quad\\
            no & no   & \antigenposNN\,\,\,\,\,\,\, & \antigennegNN\,\quad\quad\quad \\
        \bottomrule
        \end{tabular}
	};
	\node[below left=0.3cm and -2.3cm of v2] (tabPCR)
	{
		\begin{tabular}{c|rr}
        \multicolumn{1}{c}{} & \multicolumn{2}{c}{PCR Test} \\ 
        \cmidrule{2-3}
        \multicolumn{1}{c}{COVID-19} & \multicolumn{1}{c}{positive} & \multicolumn{1}{c}{negative} \\ 
        \hline\midrule
        yes & \dboxed{\textcolor{black}{\PCRposCOVIDY}}$\,\,\textcolor{blue-pigment}{q}$ & \dboxed{\textcolor{black}{\PCRnegCOVIDY}}$\,\,\textcolor{blue-pigment}{1{-}q}$\\
        no & \PCRposCOVIDN\,\,\,\,\,\,\, & \PCRnegCOVIDN \,\,\quad\quad\quad   \\
        \bottomrule
        \end{tabular}
	};
	\path (v0) edge[-{Latex[length=2mm]}] (v3)
	(v1) edge[-{Latex[length=2mm]}] (v0)
	(v1) edge[-{Latex[length=2mm]}] (v2) 
	(v1) edge[-{Latex[length=2mm]}] (v3);
	

	\end{tikzpicture}
	}
	\end{minipage}
   }
   \hfill
   \centering
  \subfloat[
  ]{
      \label{fig:pmc-example}
  \centering
  \begin{minipage}{0.2\linewidth}
  \centering
      \resizebox{0.44\width }{0.44\height}{%
      \begin{tikzpicture}
    [
		node distance=0.3cm and 0.3cm,
		mNode/.style={draw,ellipse,align=center, minimum size=0.5cm},
		mLNode/.style={align=center, minimum size=0.5cm}
		]
		\node (dummy){};
		\node[mNode,below=0.5cm of dummy] (init){init};
		\node[mNode,below left=0.7cm and 0.2cm of init] (cy) {C=yes};
		\node[mNode,below right=0.7cm and 0.2cm of init] (cn) {C=no};
		
		\node[mNode,below=0.7cm of cy] (cysn) {C=yes \\ S=no};
		\node[mNode,left=0.5cm of cysn] (cysy) {C=yes \\ S=yes};
		\node[mNode,below=0.7cm of cn] (cnsy) {C=no \\ S=yes};
		\node[mNode,right=0.5cm of cnsy] (cnsn) {C=no \\ S=no};
		
		\node[mNode,below=1cm of cysn] (syan) {C=yes \\ A=neg};
		\node[mNode,left=0.5cm of syan] (syap) {C=yes \\ A=pos};
		\node[mNode,below=1cm of cnsy] (snap) {C=no \\ A=pos};
		\node[mNode,right=0.5cm of snap] (snan) {C=no \\ A=neg};
		
		\node[mNode,below=1.8cm of syan] (pp) {P=pos};
		\node[mNode,below=1.8cm of snap] (pn) {P=neg};
		
		\draw [-{Latex[length=2mm]}] (dummy) -- (init);
		\draw [-{Latex[length=2mm]}] (init) -- (cy) node[midway, above, sloped] {\covidyes};
		\draw [-{Latex[length=2mm]}] (init) -- (cn) node[midway, above, sloped] {\covidno};
		
		\draw [-{Latex[length=2mm]}] (cy) -- (cysy) node[pos=0.5, above, sloped] {\sympyescovidyes};
		\draw [-{Latex[length=2mm]}] (cy) -- (cysn) node[pos=0.5, above, sloped] {\sympnocovidyes};
		\draw [-{Latex[length=2mm]}] (cn) -- (cnsy) node[pos=0.5, above, sloped] {\sympyescovidno};
		\draw [-{Latex[length=2mm]}] (cn) -- (cnsn) node[pos=0.5, above, sloped] {\sympnocovidno};	
		
		\draw [-{Latex[length=2mm]}] (cysy) -- (syap) node[pos=0.5, left] {\textcolor{blue-pigment}{$p$}};
		\draw [-{Latex[length=2mm]}] (cysy) -- (syan) node[pos=0.15, below, sloped] {\textcolor{blue-pigment}{$1{-}p$}};
		\draw [-{Latex[length=2mm]}] (cnsy) -- (snap) node[pos=0.5, below, sloped] {\antigenposNY};
		\draw [-{Latex[length=2mm]}] (cnsy) -- (snan) node[pos=0.2, below, sloped] {\antigennegNY};	
		\draw [-{Latex[length=2mm]}] (cysn) -- (syap) node[pos=0.25, above, sloped] {\antigenposYN};
		\draw [-{Latex[length=2mm]}] (cysn) -- (syan) node[pos=0.5, above, sloped] {\antigennegYN};
		\draw [-{Latex[length=2mm]}] (cnsn) -- (snap) node[pos=0.15, below, sloped] {\antigenposNN};
		\draw [-{Latex[length=2mm]}] (cnsn) -- (snan) node[pos=0.5, above, sloped] {\antigennegNN};	
		
		\draw [-{Latex[length=2mm]}] (syap) -- (pp) node[pos=0.2, left] {\textcolor{blue-pigment}{$q$}};
		\draw [-{Latex[length=2mm]}] (syap) -- (pn) node[pos=0.1, above, sloped] {\textcolor{blue-pigment}{$1{-}q$}};
		\draw [-{Latex[length=2mm]}] (syan) -- (pp) node[pos=0.2, left] {\textcolor{blue-pigment}{$q$}};
		\draw [-{Latex[length=2mm]}] (syan) -- (pn) node[pos=0.15, below, sloped] {\textcolor{blue-pigment}{$1{-}q$}};	
		\draw [-{Latex[length=2mm]}] (snap) -- (pp) node[pos=0.15, above, sloped] {\PCRposCOVIDN};
		\draw [-{Latex[length=2mm]}] (snap) -- (pn) node[pos=0.17, below, sloped] {\PCRnegCOVIDN};
\draw [-{Latex[length=2mm]}] (snan) -- (pp) node[pos=0.1, above, sloped] {\PCRposCOVIDN};
		\draw [-{Latex[length=2mm]}] (snan) -- (pn) node[pos=0.25, below, sloped] {\PCRnegCOVIDN};
		
		\path[every loop/.append style=-{Latex[length=2mm]}]
		(pn) edge [loop below] node {1} (pp)
		(pp) edge [loop below] node {1} (pn);
		
	\end{tikzpicture}
	}
	\end{minipage}
	}
	\hfill
	\centering
\subfloat[
  ]{ 
      \label{fig:e-tailored-pmc}
  \centering
  \begin{minipage}{0.2\linewidth}
  \centering
  \resizebox{0.6\width }{0.6\height}{%
  
    \begin{tikzpicture}
    [
		node distance=0.3cm and 0.3cm,
		mNode/.style={draw,ellipse,align=center, minimum size=0.5cm},
		mLNode/.style={align=center, minimum size=0.5cm}
		]
		\node (dummy){};
		\node[mNode,below=0.5cm of dummy] (init){$s_0$};
		\node[mNode,below left=0.9cm and 0.5cm of init] (cy) {$s_1$};
		\node[mNode,below right=0.9cm and 0.5cm of init] (cn) {$s_2$};
		
		\node[mNode,below=0.9cm of cy] (cysn) {$s_4$};
		\node[mNode,left=0.7cm of cysn] (cysy) {$s_3$};
		\node[mNode,below=0.9cm of cn] (cnsy) {$s_5$};
		\node[mNode,right=0.7cm of cnsy] (cnsn) {$s_6$};
		
		\node[mNode,below=0.9cm of cysn] (syap) {$s_7$};
		\node[mNode,below=0.9cm of cnsy] (snap) {$s_8$};
		
		\node[mNode,below=0.7cm of syap] (pp) {$s_9$};
		\node[mNode,accepting,below=0.7cm of snap] (pn) {$s_{10}$};
		
		\draw [-{Latex[length=2mm]}] (dummy) -- (init);
		\draw [-{Latex[length=2mm]}] (init) -- (cy) node[midway, above, sloped] {\covidyes};
		\draw [-{Latex[length=2mm]}] (init) -- (cn) node[midway, above, sloped] {\covidno};
		
		\draw [-{Latex[length=2mm]}] (cy) -- (cysy) node[pos=0.5, above, sloped] {\sympyescovidyes};
		\draw [-{Latex[length=2mm]}] (cy) -- (cysn) node[pos=0.5, below, sloped] {\sympnocovidyes};
		\draw [-{Latex[length=2mm]}] (cn) -- (cnsy) node[pos=0.5, above, sloped] {\sympyescovidno};
		\draw [-{Latex[length=2mm]}] (cn) -- (cnsn) node[pos=0.5, above, sloped] {\sympnocovidno};	
		
		\draw [-{Latex[length=2mm]}] (cysy) -- (syap) node[pos=0.2, below] {\textcolor{blue-pigment}{$p$}};
		
				 \draw [-{Latex[length=2mm]}] (cysy) to [bend left=30] node [above, sloped] (TextNode1) {\textcolor{blue-pigment}{$1{-}p$}} (init);
				 
		\draw [-{Latex[length=2mm]}] (cnsy) -- (snap) node[pos=0.5, above, sloped] {\antigenposNY};
		
		 \draw [-{Latex[length=2mm]}] (cnsy) to [bend left=15] node [pos=0.25,above, sloped] (TextNode1) {\antigennegNY} (init);
		 
		\draw [-{Latex[length=2mm]}] (cysn) -- (syap) node[pos=0.5, below, sloped] {\antigenposYN};
		
			 \draw [-{Latex[length=2mm]}] (cysn) to [bend right=15] node [pos=0.25,above, sloped] (TextNode1) {\antigennegYN} (init);
			 
		\draw [-{Latex[length=2mm]}] (cnsn) -- (snap) node[pos=0.3, below, sloped] {\antigenposNN};
		
	 \draw [-{Latex[length=2mm]}] (cnsn) to [bend right=30] node [above, sloped] (TextNode1) {\antigennegNN} (init);

		\draw [-{Latex[length=2mm]}] (syap) -- (pp) node[pos=0.5, left] {\textcolor{blue-pigment}{$q$}};

		\draw [-{Latex[length=2mm]}] (syap) to  [bend right=10]
node[pos=0.1, below, sloped]  (TextNode1) {\textcolor{blue-pigment}{$1{-}q$}}(init);
		\draw [-{Latex[length=2mm]}] (snap) -- (pn) node[pos=0.5, right] {\PCRposCOVIDN};
	\draw [-{Latex[length=2mm]}] (snap) to [bend left=10] node[pos=0.1, below, sloped] (TextNode1)  {\PCRnegCOVIDN} (init);
		
		\path[every loop/.append style=-{Latex[length=2mm]}]
		(pn) edge [loop below] node {1} (pp)
		(pp) edge [loop below] node {1} (pn);
		
	\end{tikzpicture}
  }
  \end{minipage}
  }
  \hfill
  \begin{minipage}{0.05\linewidth}
  
  \end{minipage}
  \vspace{-0.35cm}
     \caption{Our running example: (a) the  \textcolor{blue-pigment}{parametric} BN \texttt{COVID-19}, (b) the \textcolor{blue-pigment}{parametric} MC of \texttt{COVID-19}, for the topological order \emph{C;S;A;P}, (c)  the \textcolor{blue-pigment}{parametric} MC of \texttt{COVID-19} tailored to the evidence \emph{Antigen Test $=$ pos} and \emph{PCR Test $=$ pos}, abstracted from variable valuations.}
     \label{fig:pBN-pMC}
      \vspace*{-0.3cm}
 \end{figure*}

\paragraph{Bayesian networks.} 
Bayesian networks (BNs) \cite{pearl1988probabilistic} are probabilistic graphical models that enable succinct knowledge representation and facilitate probabilistic reasoning \cite{DBLP:books/daglib/0024906}. 
Parametric Bayesian networks (pBNs) \cite{DBLP:journals/tsmc/CastilloGH97}
extend BNs by allowing polynomials in conditional probability tables (CPTs) rather than constants.
\paragraph{Parameter synthesis on Markov models.} Parameter synthesis is to find the right values for the unknown parameters with respect to a given constraint. Various synthesis techniques have been developed for parametric Markov chains (pMCs) ranging over e.g., the gradient-based methods \cite{DBLP:conf/vmcai/HeckSJMK22}, convex optimization \cite{DBLP:conf/atva/CubuktepeJJKT18,DBLP:journals/tac/CubuktepeJJKT22}, and region verification \cite{DBLP:conf/atva/QuatmannD0JK16}. 
Recently, Salmani and Katoen [\citeyear{DBLP:conf/ecsqaru/SalmaniK21}] have proposed a translation from pBNs to pMCs that facilitates using pMC algorithms to analyze pBNs. Proceeding from this study, \emph{we tackle a different problem} \cite{DBLP:conf/uai/KwisthoutG08} for Bayesian networks.

\paragraph{Minimal-change parameter tuning.} Given a Bayesian network $B$, a hypothesis $H$, evidence $E$, $\lambda \in [0, 1]$, and a constraint of the form $\Pr(H | E) \leq \lambda$ (or $\geq  \lambda$),
\par
\begin{mdframed}[backgroundcolor=blue!5, nobreak=true, skipabove=4pt, skipbelow=0pt]
 what is a minimal change---with respect to a given measure of distance---in the probability values of (a subset of) CPT rows, such that the constraint holds?
\end{mdframed}
\noindent
We illustrate the problem with an example of testing COVID-19 that is adopted from several medical studies \cite{barreiro2021infection,nishiura2020estimation,dinnes2022rapid}. The outcome of PCR tests only depends on whether the person is infected, while the antigen tests are less likely to correctly identify COVID-19 if the infection is asymptomatic (or pre-symptomatic). Figure \ref{fig:pBN-example} depicts a Bayesian network that models such probabilistic dependencies.
In the original network, the probability of no COVID-19 given that both tests are positive, is $0.011089$. 
Assume that in an application domain, the result of such a query is now required not to exceed $0.009$: the aim is to make this constraint hold while imposing the least change in the original network with respect to a distance measure. 
This is an instance of the \emph{minimal-change parameter tuning} problem. We consider the $\epsilon-$bounded variant of the problem: for a subset of \emph{modifiable} CPT rows, are there new values within the distance of $\epsilon$ from the original probability values that make the constraint hold?
\paragraph{Main contributions. } Based on existing \emph{region verification} and \emph{region partitioning} techniques for pMCs,
\begin{itemize}
\vspace{-0.07cm}
\item we propose a practical algorithm for $\epsilon$-bounded tuning. 
More precisely, we find instantiations that (i) satisfy the constraint (if the constraint is satisfiable) and (ii) are $\epsilon$-close and (iii) lean towards the minimum distance instantiation depending on a coverage factor $0 \leq \eta \leq 1$.
\item 
We propose two region expansion schemes to realize $\epsilon$-closeness of the results both for Euclidean distance and for CD distance \cite{DBLP:journals/ijar/ChanD05}.
\item Contrary to the existing techniques that restrict to hyperplane solution spaces, we handle pBNs with multiple parameters in multiple distributions and multiple CPTs.
\end{itemize}
\vspace{-0.07cm}
\par \noindent
Our experiments on our prototypical implementation \footnote{\url{https://github.com/baharslmn/pbn-epsilon-tuning}} indicate that $\epsilon$-bounded tuning of up to 8 parameters for large networks with 100 variables is feasible. 
\vspace{-0.125cm}
\noindent
\paragraph{Paper organization.} Section 2 includes the basic notations and Sec. 3 the preliminaries on parametric Bayesian networks. Section 4 introduces parametric Markov chains and the region verification techniques thereof. Section 5 details our main contributions and Sec. 6 our experimental results. Section 7 concludes the paper with an overview of the related studies that were not mentioned before.

\section{Background}

\paragraph{Variables.}
Let $V$ be a set of $m$ random variables $v_1, \ldots, v_m$ and $D_{v_i}$ the domain of variable $v_i$. 
For $A \subseteq V$, 
$\mbox{\it Eval}(A)$ denotes the set of joint configurations for the variables in $A$.
\paragraph{Parameters.}
Let $X \,{=}\, \{x_1, \ldots, x_n\}$ be a set of $n$ real-valued parameters.
A \emph{parameter instantiation} is a function $u\,{:}\, X \,{\rightarrow}\,\mathbb{R}$ that maps each parameter to a value.
All parameters are bounded; i.e., $lb_{x_i} \,{\leq}\, u(x_i) \,{\leq}\, ub_{x_i}$ for $x_i$. 
Let $I_i \,{=}\, [lb_{x_i}, ub_{x_i}]$. 
The \emph{parameter space} $\mathcal{U} \subseteq \mathbb{R}^{n}$ of $X$ is the set of all possible values of $X$, i.e., the hyper-rectangle spanned by the intervals $I_i$ for all $i$.

\paragraph{Substitution.}
Polynomials $f$ over $X$ are functions $f: \mathbb{R}^{n} \rightarrow \mathbb{R}$ where $f(u)$ is obtained by replacing each occurrence of $x_i$ in $f$ by $u(x_i)$; e.g., for $f = 2x_1^2{+}x_2$, $u(x_1) = 3$ and $u(x_2) = 2$, $f(u) = 20$.
Let $f[u]$ denote such substitution.
\paragraph{Parametric distributions.} Let $\Distr(D)$ denote the set of probability distributions over $D$. Let $\mathbb{Q}[X]$ be the set of multivariate polynomials with rational coefficients over $X$. A \emph{parametric probability distribution} is the function $\mu: D \rightarrow \mathbb{Q}[X]$ with $\sum_{d \in D} \mu(d) \,{=}\, 1$.  Let $\mbox{\it pDistr}(D)$ denote the set of \emph{parametric} probability distributions over $D$ with parameters in $X$. Instantiation $u$ is \emph{well-formed} for $\mu \,{\in}\, \mbox{\it pDistr}(D)$ iff $0 \leq \mu(d)[u] \leq 1$ and $\Sigma_{d \in D} \, \mu(d)[u] = 1$.
\paragraph{Co-variation scheme. } 
A co-variation scheme $cov: \Distr(D) \times D \to \pDistr(D)$ maps a probability distribution $\mu$ to a parametric distribution $\mu'$ based on a given  $d \in D$.
 \paragraph{Distance measure. }
The function $d: \mathcal{U} \times \mathcal{U}\rightarrow \mathbb{R}^{\geq 0}$ is a \emph{distance measure} if for all $u, u', u'' \in \mathcal{U}$, it satisfies: (I) positiveness: $d(u,u') \geq 0$, (II) symmetry: $d(u,u') = d(u',u)$, and (III) triangle inequality: $d(u,u'') \leq d(u,u') + d(u',u'')$.

\DeclareRobustCommand{\hlcyan}[1]{{\sethlcolor{cyan}\hl{#1}}}
\section{Parametric Bayesian Networks}
\label{sec:pBN}
\noindent
A \emph{parametric Bayesian network} (pBN) is a BN in which a subset of entries in the conditional probability tables (CPTs) are polynomials over the parameters in $X$. Let $par_{v_i}$ denote the set of parents for the node $v_i$ in the graph $G$.
\begin{definition} The tuple $\mathcal{B}{=}(G, X,\Theta)$ is a \emph{parametric Bayesian network} (pBN) with directed acyclic graph $G{=}(V,W)$ over random variables $V {=} \{v_1,\ldots,v_m\}$, set of parameters $X {=}
\{x_1, \dots, x_n\}$, and \textit{parametric CPTs} $\Theta{=}\{\, \Theta_{v_i}\,{\mid}\, v_i{\in} V \, \}$ where $\Theta_{v_i} {\colon}\, \mbox{\it Eval}(par_{v_i})\,{\to}\, \pDistr(D_{v_i})$.
\label{pBN-definition}
\end{definition}

\par \noindent
The CPT row $\Theta_{v}(\overline{par})$ is a parametric distribution over $D_v$ given the parent evaluation $\overline{par}$. 
The \emph{CPT entry} $\Theta_{v}(\overline{par})(d)$,
short ${\theta}_{(v,d,\overline{par})}$, is the probability that $v{=}d$ given $\overline{par}$.
A pBN without parameters, i.e., $X = \emptyset$, is an ordinary BN.
A pBN $\B$ defines the \emph{parametric} distribution function $\Pr_{\B}$ over $\Eval(V)$. 
For well-formed instantiation $u$, BN $\B[u] = (G, \Theta[u])$ is obtained by replacing the parameter $x_i$ in the parametric functions in the CPTs of $\B$ by $u(x_i)$.
\begin{example}
Fig.~\ref{fig:pBN-example} shows a pBN over variables $V = \{C, S, A, P\}$ (initial letters of node names) and parameters $X = \{p, q\}$. 
Instantiating with $u_0(p) = \antigenposYY$ and $u_0(q) = \PCRposCOVIDY$ yields the BN $\B[u_0]$ as indicated using dashed boxes.
\end{example}

\paragraph{pBN constraints.} 
Constraints $\Pr_{B}(H|E) \sim \lambda$ involve a hypothesis $H$, an evidence $E$, $\,\sim\, \in \{\leq, \geq\}$ and threshold $0 \leq \lambda \leq 1$.
For the instantiation $u: X \rightarrow \mathbb{R}$,
\begin{equation*}
\mathcal{B}[u] \models \phi \quad \text{ if and only if } \quad \Pr_{\B[u]}(H \,|\, E) \sim \lambda.
\end{equation*}
\vspace{-0.4cm}
\paragraph{Sensitivity functions. }Sensitivity functions are rational functions that relate the result of a pBN query to the parameters \cite{DBLP:journals/tsmc/CastilloGH97,DBLP:journals/amai/CoupeG02}.
\begin{example}
\label{ex:constraint}
For the pBN $\B_1$ in Fig. \ref{fig:pBN-example}, let $\phi_1$ be the constraint:
$$
\Pr_{\B_1}(\text{C} = \text{no} \,|\,\text{A} = \text{pos} \wedge \text{P} = \text{pos} ) \leq 0.009.
$$
This reduces to the sensitivity function 
\begin{equation*}
    f_{\B_1, \phi_1} = \nicefrac{361}{34900 \cdot p \cdot q+8758 \cdot q+361}.
\end{equation*}
Using instantiation $u$ with $u(p) = 0.92075$ and $u(q) = 0.97475$, $f_{\B_1, \phi_1}[u] = 0.008798 \leq 0.009$, i.e., $\B_1[u] \models \phi_1$.
\end{example}

\vspace{0.1cm}
\paragraph{Higher degree sensitivity functions. } Contrary to the existing literature that is often restricted to multi-linear sensitivity functions, we are interested in analyzing pBNs with sensitivity functions of higher degrees. 
\begin{figure}[htb!]
	\centering
    \resizebox{0.65\width }{0.65\height}{%
	\begin{tikzpicture}[
	node distance=1cm and 0.5cm,
	mNode/.style={draw,ellipse,align=center, minimum size=0.8cm}
	]
\node[state,draw=none] (help){};
	
	\node[below=-0.5cm of help] (tabAntigen)
	{
		\begin{tabular}{cc|rr} 
            & \multicolumn{1}{c}{} & \multicolumn{2}{c}{Antigen Test 1}  \\ 
            \cmidrule{3-4}
            C& \multicolumn{1}{c}{S} & \multicolumn{1}{c}{pos} & \multicolumn{1}{c}{neg}          \\ 
            \hline\midrule
            yes  & yes    & \dboxed{\textcolor{black}{\antigenposYY}}$\,\,\textcolor{blue-pigment}{p}$ & \dboxed{\textcolor{black}{\antigennegYY}}$\,\,\textcolor{blue-pigment}{1{-}p}$  \\
            yes & no    & \dboxed{\textcolor{black}{\antigenposYN}}$\,\,\textcolor{blue-pigment}{r}$    & \dboxed{\textcolor{black}{\antigennegYN}}$\,\,\textcolor{blue-pigment}{1{-}r}$          \\
            no  & yes   & \dboxed{\textcolor{black}{\antigenposNY}}$\,\,\textcolor{blue-pigment}{s}$ & \dboxed{\textcolor{black}{\antigennegNY}}$\,\,\textcolor{blue-pigment}{1{-}s}$\\
            no & no   & \dboxed{\textcolor{black}{\antigenposNN}}$\,\,\,\textcolor{blue-pigment}{{t}}$ & \dboxed{\textcolor{black}{\antigennegNN}}$\,\,\,\,\textcolor{blue-pigment}{{1{-}t}}$ \\
        \bottomrule
        \end{tabular}
	};
	
		\node[right=0.1cm of tabAntigen] (tabAntigen2)
	{
			\begin{tabular}{cc|rr}
            & \multicolumn{1}{c}{} & \multicolumn{2}{c}{Antigen Test 2}  \\ 
            \cmidrule{3-4}
            C& \multicolumn{1}{c}{S} & \multicolumn{1}{c}{pos} & \multicolumn{1}{c}{neg}          \\ 
            \hline\midrule
            yes  & yes    & \dboxed{\textcolor{black}{\antigenposYY}}$\,\,\textcolor{blue-pigment}{p}$ & \dboxed{\textcolor{black}{\antigennegYY}}$\,\,\textcolor{blue-pigment}{1{-}p}$  \\
            yes & no    & \dboxed{\textcolor{black}{\antigenposYN}}$\,\,\textcolor{blue-pigment}{r}$    & \dboxed{\textcolor{black}{\antigennegYN}}$\,\,\textcolor{blue-pigment}{1{-}r}$          \\
            no  & yes   & \dboxed{\textcolor{black}{\antigenposNY}}$\,\,\textcolor{blue-pigment}{s}$ & \dboxed{\textcolor{black}{\antigennegNY}}$\,\,\textcolor{blue-pigment}{1{-}s}$\\
            no & no   & \dboxed{\textcolor{black}{\antigenposNN}}$\,\,\,\textcolor{blue-pigment}{{t}}$ & \dboxed{\textcolor{black}{\antigennegNN}}$\,\,\,\,\textcolor{blue-pigment}{{1{-}t}}$ \\
        \bottomrule
        \end{tabular}
	};
	
	\end{tikzpicture}
	}
    \captionsetup{justification=centering}
    \caption{Parametric CPTs for a variant of \texttt{COVID-19} example with two antigen tests; see the inter-CPT parameter dependencies.}
	\label{fig:exampleBN-dependency-cpts}
\end{figure}
\begin{example}
\label{example-pBN-depdendency}
Consider a variant of the COVID-19 example, where the person only takes the antigen test twice rather than taking both the antigen and PCR tests to diagnose COVID-19; see Fig.~\ref{fig:exampleBN-dependency-cpts} for the parametric CPTs. Then,
\begin{gather}
\scalebox{0.875}{\parbox{1.0\linewidth}{%
\begin{align*}
& f_{\B_2,\phi_2} = \Pr_{\B_2}(\text{C} = \text{no} \,|\,\text{A1} = \text{pos} \wedge \text{A2} = \text{pos} )  \\
& = \frac{950 \cdot 9 \cdot t^2 \cdot s^2}{8850 \cdot t^2 + 349 \cdot p^2 + 950 \cdot s^2 + 151 \cdot r^2 }.
\end{align*}
}}
\end{gather}
\end{example}
\noindent 

\paragraph{3.1. Parametrization.}
Let $\Theta_{\modif}$ denote the CPT entries in BN $B = (G, \Theta)$ that are explicitly changed.

\begin{definition}[BN parametrization] 
\label{def:bn:parametrization}
pBN $\B = (G, X, {\Theta}')$ is \emph{a parametrization of} BN $B= (G, \Theta)$ over $X$ w.r.t.\ $\Theta_{\modif}$ if
\begin{equation*}
    {\theta}'_{(v,d,\overline{par})} = f \text{ for some } f \in \mathbb{Q}[X] \quad \text{if} \quad  {\theta}_{(v,d,\overline{par})} \in \Theta_{\modif}.
\end{equation*}
\end{definition}

\par \noindent
The parametrization $\B$ is \emph{monotone} iff for each parametric entry \resizebox{.235\linewidth}{!}{${\theta}'_{(v,d,\overline{par})} = f$}, $f$ is a monotonic polynomial. 
The parametrization $\B$ is \emph{valid} iff $\B$ is a pBN, i.e., iff \resizebox{.35\linewidth}{!}{${\Theta}'_{v}(\overline{par}) \in \pDistr(D_{v})$} for each random variable $v$ and its parent evaluation $\overline{par}$.
To ensure validity, upon making a CPT entry parametric, the complementary entries in the row should be parametrized, too. This is ensured by \emph{co-variation} schemes. The most established co-variation scheme in the literature is the \emph{linear proportional} \cite{DBLP:journals/amai/CoupeG02} scheme that has several beneficial characteristics \cite{DBLP:journals/ijar/Renooij14}, e.g., it preserves the ratio of CPT entries.

\begin{definition}[Linear proportional co-variation] 
A \emph{linear proportional co-variation} over $X$ maps the CPT $\Theta_v$ onto the parametric CPT ${\Theta}'_v$ based on $d_k \in D_v$, where
\scalebox{0.905}{\parbox{1.0\linewidth}{%
\begin{align*}
\left\{
	\begin{array}{ll}
		 {\theta}'_{(v,d_k,\overline{par})} = x  & \mbox{for some } x \in X \\ 
		{\theta}'_{(v,d_j,\overline{par})} = \dfrac{1{-}x}{1 {-} {\theta}_{(v,d_k,\overline{par})}} \cdot {\theta}_{(v,d_j,\overline{par})}  & \mbox{for } d_j \neq d_k \in D_v.
	\end{array}
\right.
\end{align*}
}}
\label{def:proportional-covariation}
\end{definition}
\par \noindent 
Note that we allow the repetition of parameters in multiple distributions and multiple CPTs, see e.g., Fig.~\ref{fig:exampleBN-dependency-cpts}.

\paragraph{3.2. Formal Problem Statement.}
\par
\noindent
Consider BN $B=(G,\Theta)$, its valid parametrization $\B = (G, X, \Theta')$ over $X$ with constraint $\phi$. 
Let $u_0$ be the original value of the parameters $X$ in $B$ and $d : \mathcal{U}\times \mathcal{U} \to \mathbb{R}^{\geq 0}$ a distance measure. {Let $\hat{d_0}$ denote an upperbound for $d_{\{u \in \mathcal{U}\}}(u_0,u)$.}
 The \emph{minimum-distance parameter tuning problem} is to find $u_{\min} = \argmin_{\{u \in \mathcal{U}\,|\, \B[u] \models \phi\}} d(u,u_0)$.
Its generalized variant for {$0 \leq \epsilon \leq \hat{d_0}$} and $0 \leq \eta \leq 1$ is:
\label{problem-statement-pBN}
\begin{mdframed}[backgroundcolor=blue!5, nobreak=true, skipabove=4pt, skipbelow=0pt]
 The \emph{$(\epsilon, \eta)$-parameter tuning problem} is to find $u \,{\in}\, \mathcal{U}$ s.t.\ $\B[u] \models \phi$, $d(u, u_0) \,{\leq}\, \epsilon$, and {$d(u, u_{\min}) \,{\leq}\, (1{-}\eta) \cdot \,\hat{d_0}$}.
\end{mdframed}
\par \noindent 
$(\hat{d_0},1)$-tuning gives the minimum-distance tuning. We call $\eta$ the coverage factor that determines, intuitively speaking, the minimality of the result; we discuss this in Sec.~\ref{subec:minimal-change}. We consider two distance measures.
\paragraph{Euclidean distance (EC-distance).} 
The $EC$-distance between $u$ and $u'$ (both in $\mathcal{U}$) is defined as:
\scalebox{0.820}{\parbox{1.0\linewidth}{%
\begin{align*}
EC(u, u') = \sqrt{ \sum_{x_i \in X} \big(u(x_i) - u'(x_i)\big)^2}.
\end{align*}
}}
\par \noindent \textbf{Corollary 1. }
  For $n = |X|$, {$\hat{d_0} = \sqrt{n}$ is an upperbound for the EC distance of any $u_0$ from any instantiation $u \,{\in}\, \mathcal{U}$}. 
  \vspace{0.1cm}
 \par \noindent \textbf{Corollary 2. }Let $0 \leq \epsilon \leq \sqrt{n}$ and $u_0(x_i) - \frac{\epsilon}{{n}} \leq u(x_i) \leq u_0(x_i) +\frac{\epsilon}{{n}}$. Then, $EC(u,u_0) \leq \epsilon$.

\paragraph{Chan-Darwiche distance (CD distance)} 
\cite{DBLP:journals/ijar/ChanD05} is a distance measure to quantify the distance between the probability distributions of two BNs, defined as: 
\scalebox{0.820}{\parbox{1.0\linewidth}{%
\begin{align*}
\CD(u , u') =  \ln \, \underset{ w \in \Eval({V})}{\max} \frac{\Pr_{\B[u']}(w)}{\Pr_{\B[u]}(w)} \,\,\,\,\,  {-}  \ln \,  \underset{ w \in \Eval({V})}{\min} \, \frac{\Pr_{\B[u']}(w)}{\Pr_{\B[u]}(w)},
\end{align*}
}}
 \par \noindent 
where both $\nicefrac{0}{0} = 1$ and $\nicefrac{\infty}{\infty} = 1$ by definition. Whereas the EC-distance can be computed in $O(n)$, this is for CD-distances NP-complete \cite{DBLP:conf/uai/KwisthoutG08}. It has known closed-forms only for limited cases, e.g., single parameter and single-CPT pBNs \cite{DBLP:conf/uai/ChanD04}. 
Let $\Theta_v$ be the CPT that is parametrized to $\Theta'_v$ by a monotone parametrization. Let the minimum (maximum) probability entry of $\Theta_v$ be $\theta_{\min}$ ($\theta_{\max}$) parametrized to $\theta'_{\min}$ with $x$ ($\theta'_{\max}$ with $y$).
Let $lb_x,ub_x,lb_y,ub_y \neq 0,1$ be the upperbounds and the lowerbounds of $x$ and $y$ in $\mathcal{U}$.
\vspace{0.1cm}
\par \noindent \textbf{Corollary 3. }An upperbound for the CD distance of $u_0$ from any instantiation $u \,{\in}\, \mathcal{U}$ is:
\scalebox{0.875}{\parbox{1.0\linewidth}{%
\begin{align*}
\hat{d_0} =\ln\frac{\max{\{\theta'_{\min}[lb_{x}],\theta'_{\min}[ub_{x}]\}}}{\theta_{\min}} - \ln\frac{\min{\{\theta '_{\max}[lb_{y}],\theta'_{\max}[ub_{y}]\}}}{\theta_{\max}}
\end{align*}
}}
as derived from the single CPT closed-form \cite{chan2005sensitivity}.
\par \noindent \textbf{Corollary 4. }Let $0 \,{\leq}\, \epsilon\, {\leq} \,{\hat{d_0}}$ and $\alpha {=} e^{\scriptsize \nicefrac{\epsilon}{2}}$. Let for each $\theta \in \Theta_v$, $ \alpha^{-1} \,{\cdot}\, \theta'[u_0] \leq \theta'[u] \leq \alpha \,{\cdot}\, \theta'[u_0]$ . Then, {$\CD(u,u_0) \leq \epsilon$}. 
\vspace{0.15cm}
\par \noindent Note that zero probabilities are often considered to be logically impossible cases in
the BN, see e.g., \cite{DBLP:conf/kr/KisaBCD14}. Changing
the parameter values from (and to) $0$ yields CD distance $\infty$. We consider $\epsilon-$bounded parameter tuning: we (i) forbid zero probabilities
in $\Theta_{\modif}$ (the CPT entries that are explicitly modified), (ii) use the linear proportional co-variation scheme 
 (Def.~\ref{def:proportional-covariation}) that is zero-preserving
\cite{DBLP:journals/ijar/Renooij14}, i.e., it forbids changing the co-varied parameters from non-zero to zero. This co-variation scheme, in general, optimizes the CD-distance for single CPT pBNs, see \cite{DBLP:journals/ijar/Renooij14}.
\vspace{0.1cm}

\captionsetup[subfigure]{justification=centering}
 
 \begin{figure*}[!ht]
   \vspace*{-0.7cm}
     \centering
   \subfloat[The original (sub-)pMC, \\ $\textcolor{blue-pigment}{\antigenposNNMin}\leq t \leq \textcolor{color4}{\antigenposNNMax}.$ ]{
   \label{pla-a}
   \centering
     \begin{minipage}{0.3\textwidth}
     \centering
      \resizebox{0.5\width }{0.5\height}{%
          \begin{tikzpicture}[
		node distance=0.3cm and 0.3cm,
		mNode/.style={draw,circle,align=center, minimum size=0.6cm},
		mLNode/.style={align=center, minimum size=0.5cm}
		]
		\node (dummy){};
		\node[mNode,right=0.8cm of dummy] (cn) {C=no};
		
		\node[mNode,above right=0.1cm and 1.2cm of cn] (cnsy) {C=no \\ S=yes};
		\node[mNode,below right=0.1cm and 1.2cm of cn] (cnsn) {C=no \\ S=no};

		\node[mNode,above right=0.1cm and 1.2cm of cnsn] (snap) {C=no \\ A=pos};
		\node[mNode,below right=0.1cm and 1.2cm of cnsn] (snan) {C=no \\ A=neg};
		
	
	\node[mNode,right= 1.3cm of snap] (pp) {P=pos};
	\node[mNode,right=1.3cm of snan] (pn) {P=neg};
		
		\draw [-{Latex[length=2mm]}] (dummy) -- (cn) node[midway, above,sloped] {\covidno};
		
		\draw [-{Latex[length=2mm]}] (cn) -- (cnsy) node[pos=0.5, above, sloped] {\sympyescovidno};
		\draw [-{Latex[length=2mm]}] (cn) -- (cnsn) node[pos=0.5, below, sloped] {\sympnocovidno};	
		
		\draw [-{Latex[length=2mm]}] (cnsn) -- (snap) node[pos=0.5, above,sloped] {\antigenposNNPAR};
		\draw [-{Latex[length=2mm]}] (cnsn) -- (snan) node[pos=0.5, below,sloped] {\antigennegNNPAR};	
		
		\draw [-{Latex[length=2mm]}] (snap) -- (pp) node[pos=0.5, above,sloped] {$t$};
		\draw [-{Latex[length=2mm]}] (snap) -- (pn) node[pos=0.8, above,sloped] {$1{-}t$};
\draw [-{Latex[length=2mm]}] (snan) -- (pp) node[pos=0.1, above, sloped] {$t$};
		\draw [-{Latex[length=2mm]}] (snan) -- (pn) node[pos=0.5, below,sloped] {$1{-}t$};
		

		\path[every loop/.append style=-{Latex[length=2mm]}]
		(pn) edge [loop right] node {1} (pp)
		(pp) edge [loop right] node {1} (pn);
		
	\end{tikzpicture}
	}
	\end{minipage}
	\hfill
   }
   \centering
  \subfloat[No parameter dependencies, \\ $\textcolor{blue-pigment}{\antigenposNNMin}\leq t, t', t'' \leq \textcolor{color4}{\antigenposNNMax}$.
  ]{ 
  \label{pla-b}
  \centering
  \begin{minipage}{0.33\textwidth}
  \centering
      \resizebox{0.5\width }{0.5\height}{%
     \begin{tikzpicture}[
		node distance=0.3cm and 0.3cm,
		mNode/.style={draw,circle,align=center, minimum size=0.6cm},
		mLNode/.style={align=center, minimum size=0.5cm}
		]
		\node (dummy){};
		\node[mNode,right=0.8cm of dummy] (cn) {C=no};
		
		\node[mNode,above right=0.1cm and 1.2cm of cn] (cnsy) {C=no \\ S=yes};
		\node[mNode,below right=0.1cm and 1.2cm of cn] (cnsn) {C=no \\ S=no};

		\node[mNode,above right=0.1cm and 1.2cm of cnsn] (snap) {C=no \\ A=pos};
		\node[mNode,below right=0.1cm and 1.2cm of cnsn] (snan) {C=no \\ A=neg};
		
	
	\node[mNode,right= 1.3cm of snap] (pp) {P=pos};
	\node[mNode,right=1.3cm of snan] (pn) {P=neg};
		
		\draw [-{Latex[length=2mm]}] (dummy) -- (cn) node[midway, above,sloped] {\covidno};
		
		\draw [-{Latex[length=2mm]}] (cn) -- (cnsy) node[pos=0.5, above, sloped] {\sympyescovidno};
		\draw [-{Latex[length=2mm]}] (cn) -- (cnsn) node[pos=0.5, below, sloped] {\sympnocovidno};	
		
		\draw [-{Latex[length=2mm]}] (cnsn) -- (snap) node[pos=0.5, above,sloped] {\antigenposNNPAR};
		\draw [-{Latex[length=2mm]}] (cnsn) -- (snan) node[pos=0.5, below,sloped] {\antigennegNNPAR};	
		
		\draw [-{Latex[length=2mm]}] (snap) -- (pp) node[pos=0.5, above,sloped] {$t'$};
		\draw [-{Latex[length=2mm]}] (snap) -- (pn) node[pos=0.8, above,sloped] {$1{-}t'$};
\draw [-{Latex[length=2mm]}] (snan) -- (pp) node[pos=0.1, above, sloped] {$t''$};
		\draw [-{Latex[length=2mm]}] (snan) -- (pn) node[pos=0.5, below,sloped] {$1{-}t''$};
		

		\path[every loop/.append style=-{Latex[length=2mm]}]
		(pn) edge [loop right] node {1} (pp)
		(pp) edge [loop right] node {1} (pn);
		
	\end{tikzpicture}
	}
	\end{minipage}
	\hfill
	}
	      \subfloat[Non-parametric MDP. ]{
       \label{pla-c}
	      \centering
   \begin{minipage}{0.35\textwidth}
   \centering
      \resizebox{0.5\width }{0.5\height}{%
	        \begin{tikzpicture}[
		node distance=0.3cm and 0.3cm,
		mNode/.style={draw,circle,align=center, minimum size=0.6cm},
		mLNode/.style={align=center, minimum size=0.5cm}
		]
		\node[](dummy){};
		\node[mNode][right=0.7cm of dummy] (init){C=no};
		\node[mNode,state,above right=0.05cm and 1.2cm of init] (s1) {C=no\\ S=yes};
		\node[mNode,state,below right=0.05cm and 1.2cm of init] (s2) {C=no \\ S=no};
		
		\node[mNode,state,above right=0.2cm and 1.5cm of s2] (s3) {C=no \\ A=pos};
		\node[mNode,state,below right=0.1cm and 1.5cm of s2] (s4) {C=no \\ A=neg};
		\node[mNode,state,right=2.65cm of s3] (s5) {P=pos};
		\node[mNode,state,right=2.65cm of s4] (s6) {P=neg};

		\draw [-{Latex[length=2mm]}] (dummy) -- (init) node[midway, above,sloped] {\covidno};
		\draw [-{Latex[length=2mm]}] (dummy) -- (init) node[midway, above, sloped] {};
		\draw [-{Latex[length=2mm]}] (init) -- (s1) node[midway, above, sloped] {\sympyescovidno};
		\draw [-{Latex[length=2mm]}] (init) -- (s2) node[midway, above, sloped] {\sympnocovidno};

		  \path[->]          (s2)  edge        [bend left=17, color=color4,dashed]          node[midway, above,sloped]  {\antigenposNNMax} (s3);

  \path[->]          (s2)  edge       [bend right=17,color=color4,dashed]           node[midway, below,sloped]  {\antigennegNNMax} (s4);
		
  \path[->]          (s2)  edge        [bend right=25,color=blue-pigment,densely dotted,line width=0.3mm]          node[midway, above,sloped]  {\antigenposNNMin} (s3);

  \path[->]          (s2)  edge       [bend left=25,color=blue-pigment,densely dotted,line width=0.3mm]           node[midway,below,sloped]  {\antigennegNNMin} (s4);

		  \path[->]          (s4)  edge        [bend left=17,color=color4,dashed]          node[pos=0.3, above,sloped]  {\PCRposCOVIDNMax} (s5);

		  \path[->]          (s4)  edge        [bend right=20,color=color4,dashed]          node[midway, below,sloped]  {\PCRnegCOVIDNMax} (s6);
		
		  \path[->]          (s4)  edge        [bend right=17,color=blue-pigment,densely dotted,line width=0.3mm]          node[pos=0.35, above,sloped]  {\PCRposCOVIDNMin} (s5);

  \path[->]          (s4)  edge       [bend left=17,color=blue-pigment,densely dotted,line width=0.3mm]           node[pos=0.8, right,above,sloped]  {\PCRnegCOVIDNMin} (s6);

		  \path[->]          (s3)  edge        [bend left=17,color=color4,dashed]          node[pos=0.4, above,sloped]  {\PCRposCOVIDNMax} (s5);

  \path[->]          (s3)  edge       [bend left=17,color=blue-pigment,densely dotted,line width=0.3mm]           node[pos=0.5, below,sloped]  {\PCRnegCOVIDNMin} (s6);

		  \path[->]          (s3)  edge        [bend right=17,color=blue-pigment,densely dotted,line width=0.3mm]          node[pos=0.4, left,above,sloped]  {\PCRposCOVIDNMin} (s5);

  \path[->]          (s3)  edge       [bend right=40,color=color4,dashed]           node[pos=0.1, below,sloped]  {\PCRnegCOVIDNMax} (s6);
		
		\path[every loop/.append style=-{Latex[length=2mm]}]
		(s5) edge [loop right] node {1} (s5);
		\path[every loop/.append style=-{Latex[length=2mm]}]
		(s6) edge [loop right] node {1} (s6);
	\end{tikzpicture}
	}
	\end{minipage}
   }
   \vspace{-0.25cm}
     \caption{The parameter lifting steps for the \texttt{COVID-19} example with parameter dependencies (Fig.~\ref{fig:exampleBN-dependency-cpts}): (a) the original (sub-)pMC $\xrightarrow{\text{{Relaxation}}}$ (b){ pMC without parameter dependencies} $\xrightarrow{\text{{Substitution}}}$ (c) {non-parametric MDP}. Note that
  $\textcolor{black}{\PCRnegCOVIDNMax}\leq 1{-}t, 1{-}t',1{-}t'' \leq \textcolor{black}{\PCRnegCOVIDNMin}$.}
     \label{fig:example-PLA}
 \end{figure*}
\section{Parametric Markov Chains}
\label{sec:pmc}
Parametric Markov chains are an extension of Markov chains (MCs) that allow parametric polynomials as transition labels:
\begin{definition}
A \emph{parametric Markov chain} (pMC) $\mathcal{M}$ is a tuple $(S, s_{I}, X, \mathcal{P})$ where $S$ is a \emph{finite} set of states, $s_I \in S$ is the initial state, $X$ is a \emph{finite} set of \emph{real-valued} parameters, and $\mathcal{P} : S  \rightarrow \pDistr(S)$ is a transition function over the states.  
\end{definition}
\begin{example}
Figures \ref{fig:pmc-example} and \ref{fig:e-tailored-pmc} are examples of pMCs. 
\end{example}

\paragraph{Reachability constraints.}
A reachability constraint $\varphi = (T, \sim, \lambda)$ is defined over pMC $\mathcal{M} = (S, s_I, X, \mathcal{P})$ with targets $T \subseteq S$, threshold $0 \leq \lambda \leq 1$, and $\sim \ \in \{\leq, \geq\}$. 
For pMC $\mathcal{M}$ and instantiation $u: X \rightarrow \mathbb{R}$: 
\begin{align*}
   \mathcal{M}[u] \models \varphi \quad \text{if and only if} \quad \Pr_{\mathcal{M}[u]}(\lozenge T) \sim \lambda, 
\end{align*}
where $\Pr_{\mathcal{M}[u]}(\lozenge T)$ denotes the probability that MC $\mathcal{M}[u]$ reaches a state in $T$.

\begin{example}
Consider the pMC $\mathcal{M}_{\B, E}$ in Fig. \ref{fig:e-tailored-pmc}. Let \resizebox{.165\linewidth}{!}{$T = \{ s_{10}\}$}. Then, \resizebox{.7\linewidth}{!}{$\Pr_{\M_{\B,E}}(\lozenge s_{10}) = \dfrac{361}{34900 \cdot p \cdot q+8758 \cdot q+361}$}.
\end{example}
\noindent
Let $R \subseteq \mathbb{R}^{n}$ with $n = |X|$ and let
\begin{align*}
    \mathcal{M}, R \models \varphi \quad \text{if and only if} \quad \forall u \in R. \ \mathcal{M}[u] \models \varphi. 
\end{align*}
\par 
\noindent
We now define the parameter synthesis problems for pMC $\M$ and reachability constraint $\varphi$ that are relevant to our setting.
\begin{definition}[Region partitioning]
\label{def:partitioning}
Partition region $R$ into $R_{+}$, $R_{-}$, and $R_{?}$ such that:
\scalebox{0.905}{\parbox{1.0\linewidth}{%
\begin{align*}
    & R_{+} \, \subseteq \,
    \underbrace{\{u \in R \mid \mathcal{M}[u] \models \varphi\},}_{\mbox{\footnotesize satisfying instantiations}} \quad  R_{-} \, \subseteq \, 
    \underbrace{\{u \in R \mid \mathcal{M}[u] \models \neg \varphi\},}_{\mbox{\footnotesize refuting instantiations}} 
\end{align*}
}}
and $R_{?} = R \setminus (R_{+} \cup R_{-})$ with $||R_{?}|| \leq (1{-}\eta){\cdot}||R||$ for some given \emph{coverage factor} $0 \leq \eta \leq 1$. 
\end{definition}
The sub-region $R_{?}$ denotes the fragment of $R$ that is \emph{inconclusive} for $\varphi$. 
This fragment should cover at most fraction $1{-}\eta$ of $R$'s volume. 
Region partitioning is exact if $R_{?} = \emptyset$.


\begin{definition}[Region verification] For the region $R$ and the specification $\phi$, the problem is to check whether:
\scalebox{0.905}{\parbox{1.0\linewidth}{%
\begin{align*}
    & \underbrace{\mathcal{M}, R \models \varphi}_{R \text{ is accepting}} \ \text{ or } \   \underbrace{\mathcal{M}, R \models \neg \varphi}_{R \text{ is rejecting}} \  \text{ or } \ 
    \underbrace{\mathcal{M}, R \not \models \varphi \,\wedge\, \mathcal{M}, R\not \models \neg \varphi}_{R \text{ is inconclusive}}.
\end{align*}
}}
\end{definition}

\paragraph{4.1. Parameter Lifting.} The parameter lifting algorithm (PLA) \cite{DBLP:conf/atva/QuatmannD0JK16} is an abstraction technique that reduces region verification to a probabilistic model checking problem for which efficient algorithms exist \cite{DBLP:conf/lics/Katoen16}. 
It first removes parameter dependencies by making all parameters unique.
Then it considers for each parameter $x_i$ only its bounds $lb_{x_i}$ and $ub_{x_i}$ within the region $R$.
This yields a non-parametric Markov Decision Process (MDP).
\begin{definition}
A \emph{Markov decision process} (MDP) is a tuple $\mathcal{M} = (S,s_I,\mbox{\it Act},\mathcal{P})$ with a finite set $S$ of states, an initial state $s_I \in S$, a finite set $\mbox{\it Act}$ of actions, and a (partial) transition probability function $\mathcal{P} : S \times \mbox{\it Act} \rightarrow \mbox{\it Distr}(S)$.
\end{definition}
\par \noindent While resolving the non-determinism in the obtained MDP, a trade-off between maximizing or minimizing $x_i$ may occur. This occurs in particular when parameter $x_i$ repeats in the outgoing transitions of multiple states, see e.g., parameter $t$ in Fig. \ref{pla-a}. 
PLA handles the issue by a \emph{relaxation} step that introduces intermediate parameters; see e.g.,  Fig.~\ref{pla-b}, where parameter $t$ in the outgoing edges of state $C{=}no, A{=}pos$ is replaced by $t'$. 
The relaxation step yields an over-approximation of the region verification problem.
\begin{example}
Consider the pMC in Fig.~\ref{fig:example-PLA}(a) and  region $t \in [\antigenposNNMin, \antigenposNNMax]$. 
Fig.~\ref{pla-b} shows the pMC after relaxation, e.g., parameter $t$ in the outgoing transitions of state $C{=}no,A{=}pos$ is replaced by $t'$. 
Fig.~\ref{pla-c} shows the MDP obtained by substituting the parameters with their extremal values. 
Consider e.g., state $C=no,S=no$.
Its outgoing dotted (blue) edges have probability $\antigenposNNMin$ and $1{-}\antigenposNNMin$ obtained from $lb(t)$. The probabilities $\antigenposNNMax$ and $1{-}\antigenposNNMax$ of the dashed (purple) edges stem from $ub(t)$.
\end{example}
\noindent
After \emph{substitution}, region verification reduces to a simple model-checking query that over-approximates the original higher-order polynomial, e.g., Example \ref{example-pBN-depdendency}.
The region accepts $\varphi$ for the pMC if the resulting MDP satisfies $\varphi$ for all schedulers.
If all the schedulers satisfy $\neg \varphi$, the region is rejecting. If the region is neither accepting nor rejecting (i.e., inconclusive), it is partitioned into smaller subregions that are more likely to be either accepting or rejecting. 
Partitioning ends when at least $\eta\%$ of the region is conclusive.

\paragraph{4.2. Parametric Markov Chain for pBNs.}
To enable the use of PLA to parameter tuning of pBNs, we map pBNs onto pMCs as proposed in \cite{DBLP:conf/ecsqaru/SalmaniK21,DBLP:journals/corr/abs-2105-14371}; see the next example.

\begin{example} 
\label{example:pBN2pMC}
Consider the acyclic pMC in Fig.~\ref{fig:pmc-example} for the pBN in Fig. \ref{fig:pBN-example} for the topological ordering $\varrho: C < S < A < P$. 
Initially, all variables are \emph{don't care}. 
The initial state can evolve into $\mbox{\it C}{=}yes$ and $\mbox{\it C}{=}no$.
Its transition probabilities $\covidyes$ and $\covidno$ come from the CPT entries of \mbox{\it C}.
Generally, at ``level'' $i$ of the pMC, the outgoing transitions are determined by the CPT of the $i{+}1$-st variable in $\varrho$. 
\end{example}

\noindent
pBN constraints relate to reachability constraints in its pMC.
\begin{example}
Consider the pBN $\B_1$ from Fig. \ref{fig:pBN-example} and the query \resizebox{.55\linewidth}{!}{$\Pr_{\B}(\text{C} = \text{no} \,|\,\text{A} = \text{pos} \wedge \text{P} = \text{pos} )$} from Example \ref{ex:constraint}. Let $\M^{\varrho}_{\B}$ be the pMC in Fig. \ref{fig:pmc-example}. The query coincides with
\scalebox{0.905}{\parbox{1.0\linewidth}{%
\begin{align*}
& \dfrac{1 - \Pr_{\M^\varrho_\B}(\lozenge\, \text{C}=\text{yes} \lor \text{A} = \text{neg} \lor \text{P} = \text{neg} )}{1 - \Pr_{\M^\varrho_\B}(\lozenge\, \text{A} = \text{neg} \lor \text{P} = \text{neg})} \\
& = \dfrac{361}{34900 \cdot p \cdot q+8758 \cdot q+361} = f_{\B, \phi}; \text{ see Ex. \ref{ex:constraint}.}
\end{align*}
}}
\end{example}

\begin{figure*}[h]
\centering
\includegraphics[scale=0.13]{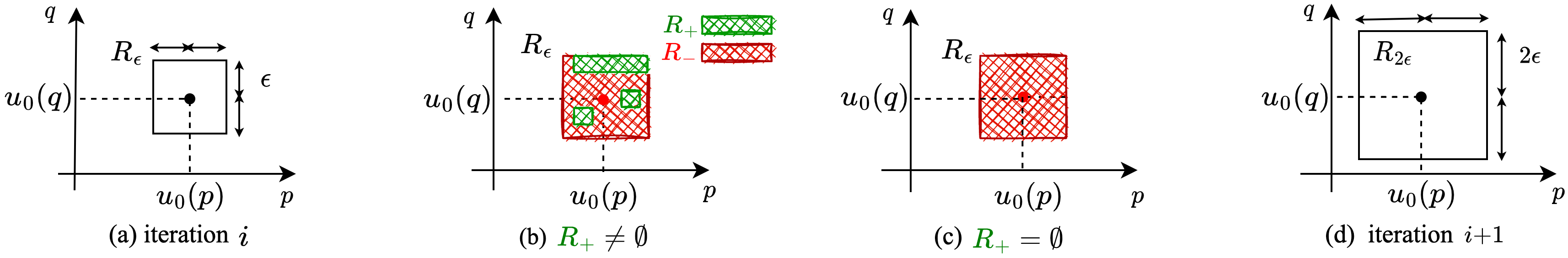}
\caption{The iterative procedure of region-based parameter tuning for $n=2$ and $\gamma = \nicefrac{1}{2}$.}
\label{fig:iterative-tuning}
\end{figure*}
\section{Region-based Minimal Change Tuning}
\label{subec:minimal-change}
Our approach to the parameter tuning problem for a pBN and constraint $\phi$ consists of two phases.
The $\epsilon$-close partitioning
exploits parameter lifting on the pMC for reachability constraint $\varphi$ of $\phi$.
We start with a rectangular region enclosing the original instantiation $u_0$.
This region is iteratively expanded if it rejects $\varphi$. 
Inconclusive regions are partitioned. 
Any iteration that finds some accepting subregion ends the first phase. 
The second phase (Alg.~\ref{alg:minimal-inst}) 
extracts a minimal-change instantiation from the set of accepting sub-regions.
\newcommand{\Cross}{\mathbin{\tikz [x=1.4ex,y=1.4ex,line width=.2ex] \draw (0,0) -- (1,1) (0,1) -- (1,0);}}%

\begin{algorithm}[!h]
\scriptsize

   $\M_{\B} \leftarrow \computeMC(\B)$
   
   $\varrho\leftarrow \reachSpec(\phi)$
   
    $\epsilon_0 \leftarrow \hat{d_0} \cdot \gamma^{K{-}1}$
    
     $\epsilon \leftarrow \epsilon_0$

     \Fn{\tuning{$\M_{\B},u_0,\varphi,\eta$}}{
     
    \While{$\epsilon \leq \hat{d_0}$}{
    
    $R_+ \leftarrow\, $\FEpPartition{$\mathcal{M_{\B}},\varphi,\epsilon,\eta$}
        
    \eIf{$R_+ = \emptyset$}{
    
     $\epsilon \leftarrow  \gamma^{-1} \cdot \epsilon$
    
    }{ 
        
       \KwRet{$\FminimalInst(R_{+},u_0)$}
    
    }
    }
    
    \KwRet{\NotFeas}
    }  
\vspace{0.15cm}
         \Fn{\FEpPartition{$\mathcal{M_{\B}},\varphi,\epsilon,\eta$}}{

  $R_{\epsilon} \leftarrow$ \expandRegionD{$u_0,\epsilon$}

    
         {$R_{+}, R_{-}, R_{?} \leftarrow \PSP(\mathcal{M_{\B}}, R_{\epsilon}, \varphi, \eta)$}
         
{ \KwRet{$R_+$}
}
}
   \vspace{0.15cm}
   
\Fn{\expandRegionEC{$u_0,\epsilon$}}{
  $R_{\epsilon} \leftarrow \Cross_{x_j \in X} \, [u_0(x_j) {-} \frac{\epsilon}{n}, u_0(x_j) {+} \frac{\epsilon}{n}]$ \\
   \KwRet{$R_{\epsilon}$}
}

   \vspace{0.15cm}

\Fn{\expandRegionCD{$u_0,\epsilon$}}{
$\alpha \leftarrow e^{\nicefrac{\epsilon}{2}}$ \\
  $R_{\epsilon} \leftarrow \Cross_{x_j \in X} \, [u_0(x_j) \cdot \alpha^{-1}, u_0(x_j) \cdot \alpha]$ \\
   \KwRet{$R_{\epsilon}$}
}
\caption{Minimal change tuning}
\label{alg:minimal-change}
\end{algorithm}

\vspace{-0.1cm}
\par \noindent \paragraph{5.1. Obtaining Epsilon-Close Subregions.} Alg.~ ~\ref{alg:minimal-change} is the main thread of our approach. 
Its inputs are the pBN $\B=(V,W,X,\Theta)$, the initial instantiation $u_0: X \to \mathbb{R}$, and the constraint $\phi$.
It starts with obtaining the pMC  $\mathcal{M}_{\B}$ of the pBN and reachability constraint $\varphi$ of $\phi$ (lines 1--2). The output is either (i) $u_0$ if $\mathcal{M}[u_0] \models \varphi$, or (ii) the instantiation $u$ with a minimal distance from $u_0$ or (iii) no instantiation if $\varphi$ is infeasible\footnote{That is, in the $\eta = 100\%$ of the checked parameter space, no satisfying instantiation has been found.}.  The hyper-parameters of the algorithm are the \emph{coverage factor} $0 < \eta < 1$, the \emph{region expansion factor} $0 < \gamma < 1$, and the \emph{maximum number of iterations} $K \in \mathbb{N}$.
The hyper-parameters $\gamma$ and $K$ steer the iterative procedure. They initially determine the distance bound $\epsilon$, see line 3. 
The bound $\epsilon$ determines region $R_{\epsilon}$, i.e., how far the region bounds deviate from $u_0$. 
PLA then verifies $R_{\epsilon}$. If $R_{\epsilon}$ is rejecting, $\epsilon$ is extended by the factor $\gamma^{-1}$ (l.\ 9).
Figure \ref{fig:iterative-tuning} visualizes the procedure for $\gamma{=}\nicefrac{1}{2}$ and $n{=}2$. 
At iteration $i$, (a) PLA is invoked on $R_{\epsilon}$ and either (b) $R_+ \neq \emptyset$ or (c) $R_+ = \emptyset$. For case (b), the iterative procedure ends and Alg.~\ref{alg:minimal-inst} is called to obtain an instantiation in $R_{+}$ that is closest to $u_0$. 
For case (c), the region is expanded by factor $\gamma^{-1}$ and passed to PLA, see (d). Note that the loop terminates when the distance bound $\epsilon$ reaches its upper bound $\hat{d_0}$ (l. 2). We refer to Sec.~\ref{sec:pBN} Corollary 1 and 3 for computing $\hat{d_0}$.
\vspace{0.1cm}
\par \noindent
\textbf{Region expansion schemes.} Region $R_{\epsilon}$ is determined by factor $\epsilon$, see line 16. The methods \expandRegionEC and \expandRegionCD detail this for each of our distance measures.
 For the EC distance (line 20), the parameters have an absolute deviation from $u_0$. We refer to Corollary 2, Sec. \ref{sec:pBN}. For CD distance (lines 23 and 24), the deviation is relative to the initial value of the parameter. We refer to Corollary 4, Sec. \ref{sec:pBN}.  Such deviation schemes ensure $\epsilon$-closeness of $R_{\epsilon}$ both for EC and CD distance, that is, all the instantiations in $R_{\epsilon}$ have at most distance $\epsilon$ from $u_0$. 
\vspace{0.1cm}
\par \noindent {Remark.}
For pBNs with a single parameter, already checked regions can be excluded in the next iterations.
Applying such a scheme to multiple parameters yet may not yield minimal-change instantiation.
To see this, let $u_0(x) = 0.15$ and $u_0(y) = 0.15$ and take $\epsilon_0 = 0.05$. 
Assume $R_{\epsilon_0} = [0.1,0.2] \times [0.1,0.2]$ is rejecting. 
Limiting the search in the next iteration to $R_{\epsilon_1} = [0.2,0.4] \times [0.2, 0.4]$ may omit the
accepting sub-regions that are at a closer distance from $u_0$, e.g. if the region $[0.1,0.2] \times [0.2, 0.4]$ includes some satisfying sub-regions. This is why we include the already-analyzed intervals starting from $u_0$ in the next iterations. However, as will be noted in our experimental results, this does not have a deteriorating effect on the performance: the unsuccessful intermediate iterations are normally very fast as they are often analyzed by a single model checking: the model checker in a single step determines that the entire region is rejecting and no partitioning is needed.

\begin{figure}[!ht]
\centering
\includegraphics[scale=0.13]{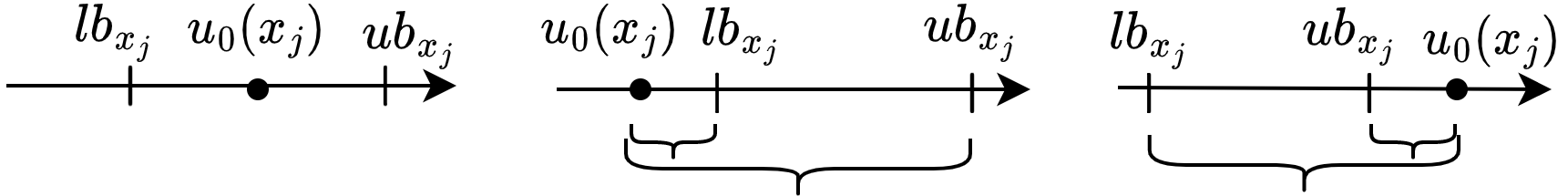}
\caption{Obtaining the minimal-change value for parameter $x \in X$.}
\label{fig:point-extraction}
\vspace*{-0.4cm}
\end{figure}

 \begin{algorithm}[!ht]
 \scriptsize
  \SetAlgoNoEnd
 \Fn{\FminimalInst{$R_+, u_0$}} 
  { 
  \tcp{$R_{+} = \{R_{+,1},\cdots,R_{+,k}\}$} 
  
  $U_{+} \leftarrow \emptyset$
  
\For{$i \leftarrow 1 \,\, \KwTo \,\, k$}{
 
  \tcp{ $R_{+,{i}} =  \Cross_{x_j \in X} [lb_{x_j}, ub_{x_j}]$}
  
  \For{$x_j \in X$}{
  \eIf{$lb_{x_j} < u_0(x_j) < ub_{x_j}$}{
  
  		$u_{+,i}(x_j) = u_0(x_j)$
  }{\eIf{$u_0(x_j) < lb_{x_j} < ub_{x_j}$}{
  
  		$u_{+,i}(x_j) = lb_{x_j}$
  }{
  
  		$u_{+,i}(x_j) = ub_{x_j}$
  }
  
	$U_{+} \leftarrow u_{+,i}$
  
  }}
  
  }
\KwRet{$\argmin\limits_{\,\,u_{+} \,\in\, U_{+}}$\,$d(u_{+},u_0)$}
}
\caption{$R_+$-minimal distance instantiation}
 \label{alg:minimal-inst}
\end{algorithm}
\paragraph{5.2. Obtaining a Minimal-Distance Instantiation.} Once $R_+ \neq \emptyset\subseteq R_{\epsilon}$ is found, it remains to find the instantiation $u_+$ in $R_{+}$ that is closest to $u_0$. 
Region $R_+$ includes infinitely many instantiations, yet minimal-distance instantiation is computable in $O(k{\cdot} n)$ for $n = |X|$ and the regions $R_{+} = \{R_{+,1}, \cdots, R_{+,k}\}$: (i) PLA ensures that the subregions $R_{+,1}$ to $R_{+,k}$ are \emph{rectangular} and \emph{mutually disjoint}. (ii) Due to the triangle equality of the distance measure, a minimum-distance instantiation can be picked from the bounded region. 
Alg.~\ref{alg:minimal-inst} simplifies finding $u_+$. 
In the first step, we pick $u_{+,i}$ from $R_{+,i}$ that has minimal distance from $u_0$ (l.\ 3--18). 
The idea is simple: for every $x_j \in X$, three cases are possible, see Fig. \ref{fig:point-extraction}. 
Let $lb_{x_j}$ be the lower bound for parameter $x_j$ in the region $R_{+,i}$ and $ub_{x_j}$ be its upper bound. 
(1) If $u_0(x_j) \in [lb_{x_j},ub_{x_j}]$, we set $u_{+,i}(x_j) = u_0(x_j)$. This yields the least distance, i.e., $0$ in the dimension of $x_j$, see Fig. \ref{fig:point-extraction} (left). 
(2) $u_0(x_j) < lb_{x_j}$, see Fig. \ref{fig:point-extraction} (middle). 
Then $lb_{x_j}$ has the least distance from $u_0(x_j)$ in the dimension $x_j$, i.e., $|lb_{x_j}{-}u_0(x_j)| < |a{-}u_0(x_j)|$ for every value $a \in [lb_x, ub_x]$  and similarly for CD-distance, $\ln{\frac{lb_{x_j}}{u_0(x_j)}} < \ln{\frac{a}{u_0(x_j)}}$.
In this case, we set $u_{+,i}(x_j) = lb_{x_j}$ to minimize the distance in dimension $x$. 
(3) By symmetry, for $u_0(x_j) > ub_{x_j}$, we set $u_{+,i}(x_j) = ub_{x_j}$; see Fig. \ref{fig:point-extraction} (right).
It remains to compute the distance for each candidate in $U_+ = \{u_{+,1}, \cdots, u_{+,k}\}$ and pick $u_+$ closest to $u_0$ (l.\ 19).


 \paragraph{Example. }
We apply the entire algorithm on the pBN in Fig. \ref{fig:pBN-example}, constraint $\phi = \Pr(\text{C}{=}\text{no} \mid \text{A}{=} \text{pos} \wedge \text{P}{=}\text{pos} ) \leq 0.009$, $u_0: p \rightarrow 0.72, q \rightarrow 0.95$, and $\eta{=}0.99$.
The pMC $\M_{\B,E}$ is given in Fig.~\ref{fig:e-tailored-pmc} and the reachability constraint is $\varphi: \Pr(\lozenge s_{10}) \leq 0.009$. 
We take $\hat{d_0} {=} \sqrt{n}$, $\gamma {=} \nicefrac{1}{2}$, and $\epsilon_0 {=} \nicefrac{1}{16}$.
\begin{enumerate}
\setlength{\itemsep}{0pt}
\item The initial region $R_{\epsilon}$ is obtained by $\epsilon{=}\nicefrac{1}{8}$: 
\begin{align*}
[0.72{-}\nicefrac{1}{8\sqrt{2}}, 0.72{+} \nicefrac{1}{8\sqrt{2}}] \times [0.95{-} \nicefrac{1}{8\sqrt{2}}, 0.95{+}\nicefrac{1}{8\sqrt{2}}].
\end{align*}
\item 
Running PLA on $R_{\epsilon}$ gives no sub-region accepting 
$\varphi$.
\item Expanding the region by the factor $\gamma^{-1} = 2$ yields:
\begin{align*}
[0.72{-}\nicefrac{1}{4\sqrt{2}}, 0.72{+} \nicefrac{1}{4\sqrt{2}}] \times [0.95{-} \nicefrac{1}{4\sqrt{2}}, 0.95{+}\nicefrac{1}{4\sqrt{2}}].
\end{align*}
\item Running PLA gives $12$ regions accepting $\varphi$, e.g., $R_{+,1}: [0.92075, 0.960375] \times [0.97475,1]$ and 
$R_{+,2}: [0.960375, 0.97028175] \times [0.9431875, 0.9495]$.
\item 
Running Alg.~\ref{alg:minimal-inst} on $R_{+,1}$ through $R_{+,12}$ gives minimal distance candidates $u_{+,1}$ through $u_{+,12}$; e.g., $u_{+,2}(p){=}lb_p{=}0.960375$ as $u_0(p) < lb_p$ and $u_{+,2}(q){=}ub_q{=}0.9495$ as $ub_q < u_0(q)$; see Fig. \ref{fig:point-extraction}.
\item 
Running Alg.~\ref{alg:minimal-inst} using the distance measure of choice---here: EC-distance---for the candidates $U_{+}$ returns $u_{+}$ closest to $u_0$: $u_+(p) = 0.92075$ and $u_+(q) = 0.97475$; distance $0.040913125$. 
\end{enumerate}

\paragraph{Discussion. } 
Let us shortly discuss the hyper-parameters: $\eta$, $\gamma$, and $K$. The hyper-parameter $\eta$ is the coverage factor for the region partitioning. Our region partitioning algorithm, i.e., parameter lifting (line 17 Alg.~\ref{alg:minimal-change}) works based on this factor: the procedure stops when $R_?$ is at most $(1-\eta) \cdot ||R||$ of the entire region $R$, e.g., for $\eta = 0.99$, it continues until at least $99\%$ of $R$ is either rejecting or accepting and at most $1\%$ is unknown. This relates $\eta$ to the approximation bounds of our algorithm, see Problem statement \ref{problem-statement-pBN}. Intuitively speaking, the coverage factor $\eta$ means that if the algorithm provides a minimal-distance value $D(u,u_0) = d$, then with the probability $1-\eta$ there may exist a smaller value distance than $d$ that works too but has not been found. One can thus consider $\eta$ as the confidence factor for the minimality of the results. The hyper-parameter $\gamma$ specifies the factor by which the region is expanded at each iteration, see Line 9, Alg.~\ref{alg:minimal-change}. The hyper-parameters $\gamma$ and $K$ specify the size of the initial region, see Lines 3 and 16 of Alg.~\ref{alg:minimal-change}. Our experiments (not detailed in the paper) with $\gamma=\{0.2, 0.5, 0.8\}$ and $K=\{2, \cdots, 12\}$ reveal that $\gamma=\nicefrac{1}{2}$ and $K = 6$ gave the best balance. Large values for $\gamma \, (0.8)$ and small $K$ lead to unnecessary computations in the initial iteration for the simple cases i.e., when small perturbations of the parameters make the constraint satisfied. Small values for $\gamma\, (0.2)$ lead to large regions in the next iterations due to the expansion by $\gamma^{-1}$.

 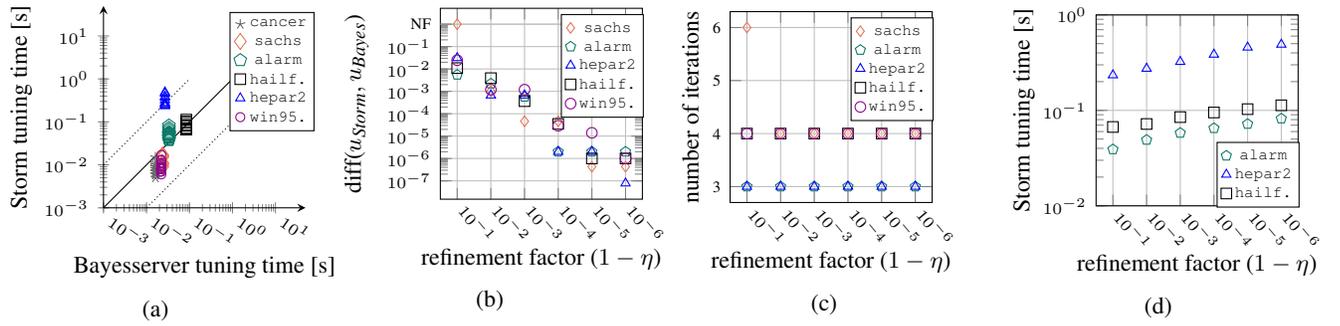
\begin{figure*}[!h]
\vspace*{-0.5cm}
     \centering
      \resizebox{1.0\width }{1.0\height}{
       \begin{minipage}{0.001\linewidth}
  \end{minipage}
  \hfill
   \subfloat[]{
   \centering
     \begin{minipage}{0.24\linewidth}
     \centering
 \scalebox{1.0}[1.0]{
        \begin{tikzpicture}
 \label{fig:storm-bayesserver-time}
    	\begin{axis}[
    		width=1.0\textwidth,
    		height=0.8\scatterplotsize,
    		axis equal image,
    		xmin=0.001,
    		ymin=0.001,
    		ymax=50,
    		xmax=50,
    		xmode=log,
    		ymode=log,
    		axis x line=bottom,
    		axis y line=left,
     		xtick={0.001,0.01,0.1,1,10},
    		xticklabels={$10^{-3}$,$10^{-2}$,$10^{-1}$,$10^{0}$,$10^1$},
    		extra x tick style = {grid = major},
     		ytick={0.001,0.01,0.1,1,10},
    		yticklabels={$10^{-3}$,$10^{-2}$,$10^{-1}$,$10^{0}$,$10^{1}$},
    		extra y tick style = {grid = major},
    		xlabel={Bayesserver tuning time [s]},
    		xlabel style={yshift=0pt},
    		ylabel={Storm tuning time [s]},
    y label style={at={(axis description cs:0.15,.5)},anchor=south},
    		yticklabel style={font=\tiny},
    		xticklabel style={rotate=325,anchor=west,font=\tiny},
    		label style={font=\small},
    		legend pos=north east,
    		legend columns=1,
    		legend style={nodes={scale=0.6, transform shape},inner sep=1.5pt,yshift=0.0cm,xshift=0.2cm},
    		]
    		\addplot[
    		scatter,
    		only marks,
    		scatter/classes={
    		    cancer={mark=star, mark size=2, color1},
    		   sachs={mark=diamond, mark size=3, dark-orange},
    			alarm={mark=pentagon, mark size=2.5, color3},
    			hailfinder={mark=square, mark size=2, black},
    			hepar2={mark=triangle, mark size=2, blue},
    			win95pts={mark=o, mark size=1.7, color4}
    		},
    		scatter src=explicit symbolic
    		]%
    		table [col sep=comma,x=p1bayesservertuningtime,y=stormtotaltuningime,meta=networkname] {plots/data/full-p1-tuning-results.csv};
    		\legend{\texttt{cancer},\texttt{sachs}, \texttt{alarm}, \texttt{hailf.}, \texttt{hepar2},\texttt{win95.}}
    		\addplot[no marks] coordinates {(0.001,0.001) (1,1)};
    		\addplot[no marks, densely dotted] coordinates {(0.001,0.01) (0.1,1)};
    		\addplot[no marks, densely dotted] coordinates {(0.01,0.001) (1,0.1)};
    	\end{axis}
\end{tikzpicture}
	}
	\end{minipage}
   }
   \hfill
   \centering
  \subfloat[
  ]{ 
  \centering
  \begin{minipage}{0.24\linewidth}
  \centering
 \scalebox{1.0}[1.0]{
  \begin{tikzpicture}
\label{fig:coverage-bayesdiff}
\begin{loglogaxis}[
    		width=1.0\textwidth,
    		height=0.7\scatterplotsize,
  xlabel = {refinement factor ($1 - \eta$)},
  ylabel = {diff($u_{\storm},u_{\bayesserver}$)},
  label style={font=\small},
      		  x dir=reverse,
  extra y ticks = {1},
extra y tick labels = {NF},
 extra y tick style = {grid = major},
  grid = major,
   xtick={0.1,0.01,0.001,0.0001,0.00001,0.000001,0.0000001,
   0.00000001,0.000000001,0.0000000001,
   0.00000000001},    
  ytick={0.0000001,0.000001,0.00001,0.0001,0.001,0.01,0.1,1},
  yticklabels={$10^{-7}$,$10^{-6}$,$10^{-5}$,$10^{-4}$,$10^{-3}$,$10^{-2}$,$10^{-1}$},
  xticklabel style={rotate=310,anchor=west,font=\tiny},
  yticklabel style={font=\tiny},
      		xlabel style={yshift=0pt},
  y label style={at={(axis description cs:0.15,.5)},anchor=south},
legend pos=north east,
    		legend columns=1,
    		legend style={/tikz/every even column/.append style={column sep=-0.2cm}, nodes={scale=0.6, transform shape},inner sep=1.5pt,yshift=0.0cm,xshift=0.1cm},
legend entries = {\texttt{sachs},\texttt{alarm},\texttt{hepar2},\texttt{hailf.},\texttt{win95.}}]
    	\addplot
    		[only marks,
    		color=dark-orange,
                mark=diamond,
                ]
    		 table [col sep=comma,x=refinementFactor,y=stormbayesserverdistancediff,meta=networkname] {plots/data/coverage-data/sachs-p1-coverage.csv};
  		    \addplot [only marks,
                  color=color3,
            mark=pentagon,
             ] table [col sep=comma,x=refinementFactor,y=stormbayesserverdistancediff,meta=networkname] {plots/data/coverage-data/alarm-p1-coverage.csv}; 
  		    \addplot [only marks,
                color=blue,
                mark=triangle,
                ] table [col sep=comma,x=refinementFactor,y=stormbayesserverdistancediff,meta=networkname] {plots/data/coverage-data/hepar2-p1-coverage.csv}; 
    		\addplot
    		[only marks,
    		color=black,
                mark=square,
                ]
    		 table [col sep=comma,x=refinementFactor,y=stormbayesserverdistancediff,meta=networkname] {plots/data/coverage-data/hailfinder-p1-coverage.csv};
    		     		\addplot
    		[only marks,
    		color=color4,
                mark=o,
                ]
    		 table [col sep=comma,x=refinementFactor,y=stormbayesserverdistancediff,meta=networkname] {plots/data/coverage-data/win95pts-p1-coverage.csv};
    		 
    		\end{loglogaxis}
\end{tikzpicture}
	}
	\end{minipage}
	}
	\hfill
	\centering
\subfloat[
  ]{ 
  \centering
  \begin{minipage}{0.24\linewidth}
  \centering
 \scalebox{1.0}[1.0]{
  \begin{tikzpicture}
\label{fig:coverage-iteration}
\begin{axis}[
    		width=1.0\textwidth,
    		height=0.7\scatterplotsize,
    		  label style={font=\small},
    		  xmode = log,
    		  x dir = reverse,
  xlabel = {refinement factor ($1 - \eta$)},
  ylabel = {number of iterations},
  	ylabel shift = -3 pt,
    xlabel style={yshift=0pt},
    y label style={at={(axis description cs:0.35,.5)},anchor=south},
 extra y tick style = {grid = major},
  grid = major,
   xtick={0.1,0.01,0.001,0.0001,0.00001,0.000001,0.0000001,
   0.00000001,0.000000001,0.0000000001,
   0.00000000001}, 
ytick={0,1,2,3,4,5,6},
yticklabels={0,1,2,3,4,5,6},
xticklabel style={rotate=310,anchor=west,font=\tiny},
yticklabel style={font=\tiny},
legend pos=north east,
legend style={nodes={scale=0.6, transform shape},inner sep=1.5pt,yshift=0.0cm,xshift=0.1cm},
legend entries = {\texttt{sachs},\texttt{alarm},\texttt{hepar2}, \texttt{hailf.},\texttt{win95.}}]
    		  		\addplot
    		[only marks,
    		color=dark-orange,
                mark=diamond,
                ]
    		 table [col sep=comma,x=refinementFactor,y=stormiteration,meta=networkname] {plots/data/coverage-data/sachs-p1-coverage.csv}; 
    		  	 \addplot [only marks,
    		  	 color=color3,
                mark=pentagon,
                ] table [col sep=comma,x=refinementFactor,y=stormiteration,meta=networkname] {plots/data/coverage-data/alarm-p1-coverage.csv}; 
  		    \addplot [only marks,
                color=blue,
                mark=triangle,
                ] table [col sep=comma,x=refinementFactor,y=stormiteration,meta=networkname] {plots/data/coverage-data/hepar2-p1-coverage.csv}; 
    		\addplot
    		[only marks,
    		color=black,
                mark=square,
                ]
    		 table [col sep=comma,x=refinementFactor,y=stormiteration,meta=networkname] {plots/data/coverage-data/hailfinder-p1-coverage.csv};
    		    		\addplot
    		[only marks,
    		color=color4,
                mark=o,
                ]
    		 table [col sep=comma,x=refinementFactor,y=stormiteration,meta=networkname] {plots/data/coverage-data/win95pts-p1-coverage.csv};

    		\end{axis}
\end{tikzpicture}
  }
  \end{minipage}
  }
  	\hfill
	\centering
\subfloat[
  ]{ 
  \centering
  \begin{minipage}{0.24\linewidth}
  \centering
 \scalebox{1.0}[1.0]{
  \begin{tikzpicture}
\label{fig:coverage-time}
\begin{loglogaxis}[
    		width=1.0\linewidth,
    		height=0.7\scatterplotsize,
    		 label style={font=\small},
    		  x dir=reverse,
    		  ymin = 0.01,
    		  ymax = 1,
  xlabel = {refinement factor ($1 - \eta$)},
  ylabel = {Storm tuning time [s]},
      		xlabel style={yshift=0pt},
    y label style={at={(axis description cs:0.2,.5)},anchor=south},
  extra y ticks = {10000},
 extra y tick labels = {NF},
 extra y tick style = {grid = major},
  grid = major,
   xtick={0.1,0.01,0.001,0.0001,0.00001,0.000001,0.0000001,
   0.00000001,0.000000001,0.0000000001,0.00000000001},  
		ytick={0.001,0.01,0.1,1},
    		yticklabels={$10^{-3}$,$10^{-2}$,$10^{-1}$,$10^{0}$},
xticklabel style={rotate=310,anchor=west,font=\tiny},
yticklabel style={font=\tiny},
legend pos=south east,
legend style={nodes={scale=0.6, transform shape},inner sep=1.5pt,yshift=-0.1cm,xshift=0.1cm},
legend entries = {\texttt{alarm},\texttt{hepar2}, \texttt{hailf.},\texttt{sachs},\texttt{win95pts}}]
    		  	 \addplot [only marks,
                color=color3,
                mark=pentagon,
                ] table [col sep=comma,x=refinementFactor,y=stormtotaltuningime,meta=networkname] {plots/data/coverage-data/alarm-p1-coverage.csv}; 
  		    \addplot [only marks,
                color=blue,
                mark=triangle,
                ] table [col sep=comma,x=refinementFactor,y=stormtotaltuningime,meta=networkname] {plots/data/coverage-data/hepar2-p1-coverage.csv}; 
    		\addplot
    		[only marks,
    		color=black,
                mark=square,
                ]
    		 table [col sep=comma,x=refinementFactor,y=stormtotaltuningime,meta=networkname] {plots/data/coverage-data/hailfinder-p1-coverage.csv};
    		 
    		\end{loglogaxis}
\end{tikzpicture}
  }
  \end{minipage}
  }
         \begin{minipage}{0.001\linewidth}
  \end{minipage}
  }
  \vspace{-0.35cm}
     \caption{The plots (a), (b): Storm vs. Bayesserver, the tuning time and the instantiation closeness; see {RQ1}. The plots (b), (c), (d): the effect of the coverage factor ($\eta$) on the tightness of the distance, the number of iterations, and the tuning time; see {RQ2}. Corner case: the number of iterations only changes when a satisfying instantiation was \emph{not found} (NF) with a certain coverage; see \texttt{sachs} at $\eta {=} 0.9$ and $\eta{=}0.99$.}
     
     \label{fig:refinement-experiments}
 \end{figure*}

 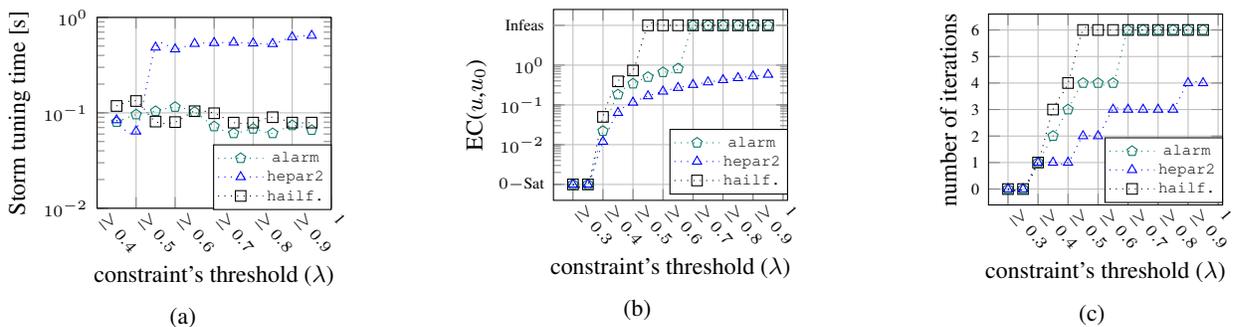
\begin{figure*}[!b]
 \vspace*{-0.7cm}
     \centering
      \resizebox{1.0\width }{1.0\height}{
       \begin{minipage}{0.1\linewidth}
  \end{minipage}
  \hfill
   \subfloat[]{
   \centering
     \begin{minipage}{0.33\linewidth}
     \centering
 \scalebox{1.0}[1.0]{
       \begin{tikzpicture}
\label{fig:threshhold-time}
\begin{axis}[
    		width=0.8\linewidth,
    		height=0.7\scatterplotsize,
    		ymode = log,
    		  label style={font=\small},
  xlabel = {constraint's threshold ($\lambda$) },
  ylabel = {Storm tuning time [s]},
      		xlabel style={yshift=-3pt},
    y label style={at={(axis description cs:0.15,.5)},anchor=south},
 extra y tick style = {grid = major},
  grid = major,
      		  ymin = 0.01,
    		  ymax = 1,
  xtick={0.1,0.2,0.3,0.4,0.5,0.6,0.7,0.8,0.9,1},
 xticklabels={$\geq 0.1$, $\geq 0.2$,$\geq 0.3$,$\geq 0.4$,$\geq 0.5$,$\geq 0.6$,$\geq 0.7$,$\geq 0.8$,$\geq 0.9$,1},  		ytick={0.001,0.01,0.1,1},
    		yticklabels={$10^{-3}$,$10^{-2}$,$10^{-1}$,$10^{0}$},
xticklabel style={rotate=310,anchor=west,font=\tiny},
yticklabel style={font=\tiny},
legend pos=south east,
legend style={nodes={scale=0.6, transform shape},inner sep=1.5pt,yshift=-0.1cm,xshift=0.1cm},
legend entries = {\texttt{alarm},\texttt{hepar2}, \texttt{hailf.},\texttt{win95pts}}]
    		  	 \addplot [only marks, 
                color=color3,
                dotted,
                mark=pentagon,
                mark options={solid},
                smooth
                ] table [col sep=comma,x=thresshold,y=stormtotaltuningime,meta=networkname] {plots/data/threshhold-data/alarm-p1-thresh.csv}; 
  		    \addplot [only marks, 
                color=blue,
                dotted,
                mark=triangle,
                mark options={solid},
                smooth
                ] table [col sep=comma,x=thresshold,y=stormtotaltuningime,meta=networkname] {plots/data/threshhold-data/hepar2-p1-thresh.csv}; 
    		\addplot
    		[only marks, color=black,
                dotted,
                mark=square,
                mark options={solid},
                smooth
                ]
    		 table [col sep=comma,x=thresshold,y=stormtotaltuningime,meta=networkname] {plots/data/threshhold-data/hailfinder-p1-thresh.csv};
    		\end{axis}
\end{tikzpicture}
	}
	\end{minipage}
   }
   \hfill
   \centering
  \subfloat[
  ]{ 
  \centering
  \begin{minipage}{0.33\linewidth}
  \centering
 \scalebox{1.0}[1.0]{
  \begin{tikzpicture}
\label{fig:threshhold-distance}
\begin{axis}[
    		width=0.8\linewidth,
    		height=0.7\scatterplotsize,
    		ymode = log,
    		  label style={font=\small},
  xlabel = {constraint's threshold ($\lambda$)},
  ylabel = {EC($u$,$u_0$)},
      		xlabel style={yshift=-3pt},
  y label style={at={(axis description cs:0.15,.5)},anchor=south},
  extra y ticks = {10},
 extra y tick labels = {Infeas},
 extra y tick style = {grid = major},
  grid = major,
  xtick={0.1,0.2,0.3,0.4,0.5,0.6,0.7,0.8,0.9,1},
 xticklabels={$\geq 0.1$, $\geq 0.2$,$\geq 0.3$,$\geq 0.4$,$\geq 0.5$,$\geq 0.6$,$\geq 0.7$,$\geq 0.8$,$\geq 0.9$,1},
  ytick={0.001,0.01,0.1,1},
  yticklabels={$0{-}$Sat,$10^{-2}$,$10^{-1}$,$10^{0}$},
xticklabel style={rotate=310,anchor=west,font=\tiny},
yticklabel style={font=\tiny},
legend pos=south east,
legend style={nodes={scale=0.6, transform shape},inner sep=1.5pt,yshift=0cm,xshift=0.1cm},
legend entries = {\texttt{alarm},\texttt{hepar2}, \texttt{hailf.},\texttt{sachs},\texttt{win95pts}}]
    		  	 \addplot [
                color=color3,
                dotted,
                mark=pentagon,
                mark options={solid},
                smooth
                ] table [col sep=comma,x=thresshold,y=stormECDistance,meta=networkname] {plots/data/threshhold-data/alarm-p1-thresh.csv}; 
  		    \addplot [
                color=blue,
                dotted,
                mark=triangle,
                mark options={solid},
                smooth
                ] table [col sep=comma,x=thresshold,y=stormECDistance,meta=networkname] {plots/data/threshhold-data/hepar2-p1-thresh.csv}; 
    		\addplot
    		[color=black,
                dotted,
                mark=square,
                mark options={solid},
                smooth
                ]
    		 table [col sep=comma,x=thresshold,y=stormECDistance,meta=networkname] {plots/data/threshhold-data/hailfinder-p1-thresh.csv};
    		\end{axis}
\end{tikzpicture}
	}
	\end{minipage}
	}
  	\hfill
	\centering
\subfloat[
  ]{ 
  \centering
  \begin{minipage}{0.33\linewidth}
  \centering
 \scalebox{1.0}[1.0]{
  \begin{tikzpicture}
\label{fig:threshhold-iteration}
\begin{axis}[
    		width=0.8\linewidth,
    		height=0.7\scatterplotsize,
    		  label style={font=\small},
  xlabel = {constraint's threshold ($\lambda$)},
  ylabel = {number of iterations},
	xlabel style={yshift=-3pt},
    y label style={at={(axis description cs:0.3,.5)},anchor=south},
 extra y tick style = {grid = major},
  grid = major,
  xtick={0.1,0.2,0.3,0.4,0.5,0.6,0.7,0.8,0.9,1},
 xticklabels={$\geq 0.1$, $\geq 0.2$,$\geq 0.3$,$\geq 0.4$,$\geq 0.5$,$\geq 0.6$,$\geq 0.7$,$\geq 0.8$,$\geq 0.9$,1},
ytick={0,1,2,3,4,5,6},
yticklabels={0,1,2,3,4,5,6},
xticklabel style={rotate=310,anchor=west,font=\tiny},
yticklabel style={font=\tiny},
legend pos=south east,
legend style={nodes={scale=0.6, transform shape},inner sep=1.5pt,yshift=-0.05cm,xshift=0.1cm},
legend entries = {\texttt{alarm},\texttt{hepar2}, \texttt{hailf.},\texttt{win95pts}}]
    		  	 \addplot [
                color=color3,
                dotted,
                mark=pentagon,
                mark options={solid},
                smooth
                ] table [col sep=comma,x=thresshold,y=stormiteration,meta=networkname] {plots/data/threshhold-data/alarm-p1-thresh.csv}; 
  		    \addplot [
                color=blue,
                dotted,
                mark=triangle,
                mark options={solid},
                smooth
                ] table [col sep=comma,x=thresshold,y=stormiteration,meta=networkname] {plots/data/threshhold-data/hepar2-p1-thresh.csv}; 
    		\addplot
    		[color=black,
                dotted,
                mark=square,
                mark options={solid},
                smooth
                ]
    		 table [col sep=comma,x=thresshold,y=stormiteration,meta=networkname] {plots/data/threshhold-data/hailfinder-p1-thresh.csv};
    		\end{axis}
\end{tikzpicture}
  }
  \end{minipage}
  }
         \begin{minipage}{0.1\linewidth}
  \end{minipage}
  }
  \vspace{-0.3cm}
     \caption{The effect of the constraint restrictiveness (the value of the threshold) on $\epsilon$-bounded tuning: (a) on the tuning time, (b) on the distance, (c) and on the number of iterations; see {RQ3}. Corner cases: for the constraints with thresholds before $0.4$, the original BN satisfies the constraints. For \texttt{hailfinder} with the thresholds $\lambda {\geq} 0.55$ and \texttt{alarm} with the thresholds $\lambda {\geq} 0.7$, the constraints become infeasible.} 
     \label{fig:threshhold-tuning-experiments}
   \vspace*{-0.4cm}
 \end{figure*}

\section{Experimental Evaluation}
We empirically evaluated our approach using a prototypical realization on top of the probabilistic model checker Storm \cite{DBLP:journals/sttt/HenselJKQV22} (version 1.7.0).  
As baseline, we used the latest version (10.4) of \emph{Bayesserver}\footnote{\url{https://www.bayesserver.com/}}, a commercial BN analysis tool that features sensitivity analysis and parameter tuning for pBNs. 
It supports pBNs with a single parameter only.
We parametrized benchmarks from \texttt{bnlearn} repository and defined different constraints. 
We (i) parametrized the CPTs of the parents (and grandparents) of the evidence nodes, and (ii) used the \emph{SamIam} tool \footnote{\url{http://reasoning.cs.ucla.edu/samiam/}}
to pick the CPT entries $\Theta_{\modif}$ most relevant to the constraint. 
To get well-formed BNs, we used the linear proportional co-variation, see Def.~\ref{def:proportional-covariation}. 
In all experiments, we picked evidences from the last nodes in the topological order; they have a long dependency path to the BN roots. 
This selection reflects the worst-case in \cite{DBLP:conf/qest/SalmaniK20}.
 We took $\gamma{=}\nicefrac{1}{2}$ and $K{=}6$ for our experiments, see Alg.\ref{alg:minimal-change}.  We conducted all our experiments on a 2.3 GHz Intel Core i5 processor with 16 GB RAM. 
 
\paragraph{RQ1: Comparison to Bayesserver.} 
We used small, medium, and large BNs from the \texttt{bnlearn} repository. 
For each BN, we parameterized one distribution in a single CPT to align with the restrictions from Bayesserver. 
Figure~\ref{fig:storm-bayesserver-time} indicates the results, comparing the tuning times (in sec) of Bayesserver (x-axis) to those of our implementation (y-axis).
The latter includes the time for region refinements by PLA. 
For each pair (pBN, constraint), we did experiments for coverage factors $\eta = 1{-}10^{-i}  \text{ for } i \in \{1, \cdots, 6\}$. 
The influence of $\eta$ on the exactness of our results is addressed under RQ2.

 \par \noindent \textbf{Findings:} 
 Storm outperforms Bayesserver for several benchmarks (\texttt{cancer}, \texttt{sachs}, \texttt{win95pts}, and multiple instances of \texttt{hailfinder}), whereas Bayesserver is faster by about an order of magnitude for the other pBNs, such as \texttt{hepar2}. \textbf{Explanation:} Bayesserver exploits specific methods for {one-way sensitivity analysis} and relies on the linear form of the sensitivity function. These techniques are very efficient yet not applicable to pBNs with multiple parameters in multiple CPTs. For our experiments on those subclasses, see {RQ4}. Such subclasses are not supported by Bayesserver and---to the best of our knowledge---not any existing BN tool. This applies e.g., also to Bayesfusion which considers only the change of a single parameter at a time, and SamIam which is limited to the single parameter and single-CPT.

\paragraph{RQ2: Sensitivity to the coverage factor $\eta$.} 
For each pBN and constraint, we decreased (the refinement factor) $1{-}\eta$ in a step-wise manner by a factor $10^{-1}$. 
To quantify the tightness of our results, we measured how our approximately-close instantiation, denoted $u_{\storm}$, differs from the absolute minimum-distance from Bayesserver, denoted $u_{\bayesserver}$. 
Figure~\ref{fig:coverage-bayesdiff} (log-log scale) plots the tightness of the results $|u_{\storm} - u_{\bayesserver}|$ (y-axis) against the refinement factor (x-axis). 
Figures~\ref{fig:coverage-iteration} and \ref{fig:coverage-time} indicate the number of iterations and the tuning time (log scale, seconds) for each refinement factor.
\par \noindent \textbf{Findings:} 
(I) Mostly, $1{-}\eta$ bounds the difference between our approximately-close solution and Bayesserver's solution.
For e.g., $\eta = 1 {-} 10^{-4}$, the difference is at most $10^{-4}$. 
(II) On increasing the coverage, the difference to the true minimal distance rapidly decreases.
((III) The computation time moderately increases on increasing the coverage, but the number of iterations was mostly unaffected. \textbf{Explanation:} (I, II) Recall that the value of $1{-}\eta$ bounds the size of the unknown regions; see Def. \ref{def:partitioning}. This indicates why $|u_{\storm} - u_{\bayesserver}|$ relates to $1{-}\eta$. (III) At a higher coverage factor, the region partitioning is more fine-granular possibly yielding more accepting regions to analyze. Therefore the computation becomes more expensive. The timing is, however, not correlated to the number of iterations. This is because the $\epsilon$-close iterations before the last iteration are often completed by a single region verification and are very fast. Similar observations have been made for $n{>}1$ parameters; see RQ4.
  
\begin{table*}[!h]
\vspace*{-0.2cm}
\begin{tabular}{ll}
		\scalebox{0.80}{
			\begin{tabular}{ccrrrrr}
		\toprule
		\multicolumn{2}{c}{pBN info} & \multicolumn{1}{c}{constraint} & setting & \multicolumn{3}{c}{results} \\ 
		 \cmidrule(lr){1-2} \cmidrule(lr){3-3}
\cmidrule(lr){4-4} \cmidrule(lr){5-7} 
 pCPT   & par & thresh. & cover. & EC & iter & \textbf{t}(s)  \\ \midrule
		         $1$       & 2 & $\geq 0.115$ & 99\%   &  0.8365285878 & 5 & 4.200 \\
				 $1$       & 2 & $\geq 0.25$  & 100\%   &  Infeasible &6 & 3.750 \\
				 $3$       & 4  & $\geq 0.115$  & 90\%   &  0.1350477346 & 2 & 5.210 \\
		   		 $3$       & 4 & $\geq 0.25$  & 80\%   &  0.4479118217 & 4 & 5.210 \\
				 $3$       & 4  & $\geq 0.25$  & 90\%   &  0.4238956239 & 4 & 30.82 \\
		        $3$       & 4 & $\geq 0.25$  & 95\%   &  0.4098399078 & 4 & 1128.5 \\
			     $3$       & 4 & $\geq 0.25$ &  99\%   &  - & - & TO \\
				 $4$       & 8  & $\geq 0.115$  & 80\%   &  0.8805679985 & 2 & 4.496 \\
			     $4$       & 8 & $\geq 0.115$  & 90\%   &  0.1350477346 & 2 & 15.91\\
		        $4$       & 8  & $\geq 0.30$ & 60\%   &  0.8805679985 & 6 & 809.7 \\
				   $4$       & 8  & $\geq 0.35$  & 99.99\%   &  Infeasible & 6 & 3.757 \\
       	     $4$       & $8^*(2)(2)$ & $\geq 0.115$  & 80\%   &  0.013153540 & 1 & 180.6 \\
			        $4$       & 16 & $\geq 0.35$  & 10\%   &  - & 1 & MO \\
	        \bottomrule
			\end{tabular}
	}

&

	\scalebox{0.80}{
			\begin{tabular}{ccrrrrr}
		\toprule
		\multicolumn{2}{c}{pBN info} & \multicolumn{1}{c}{constraint} & setting & \multicolumn{3}{c}{results} \\ 
		 \cmidrule(lr){1-2} \cmidrule(lr){3-3}
\cmidrule(lr){4-4} \cmidrule(lr){5-7} 
 pCPT   & par & thresh. & cover. & EC & iter & \textbf{t}(s)  \\ \midrule
	 $1$       & 2  & $\leq 0.30$   & 85\%    &  1.138343094  & 5 & 2.090 \\
$1$       & 2  & $\leq 0.30$  & 99\%    &  1.091371208   & 5 & 2.134 \\
 $1$       & 2  & $\leq 0.30$  & 99.99\%    &  1.090202067    & 5 & 302.1 \\
 $1$       & 4  & $\leq 0.30$   & 70\%    & 0.909491616 &  5 & 5.701 \\
	 $1$       & 4  & $\leq 0.30$   & 85\%    & 0.890217670 & 5 & 142.1 \\
$1$       & 4  & $\leq 0.30$   & 90\%    & - & -  & TO\\
	$1$       & 4  & $\leq 0.20$  & 100\%    & Infeasible &  6 & 1.869 \\
 	$1$       & $4^* (3)$ & $\leq 0.30$  & 99\%    & 0.472406250 &  5 & 2.731 \\
			 $1$       & 8  & $\leq 0.30$ & 20\%    & 0.5628159113 & 5 & 534.6  \\
		$1$       & 8  & $\leq 0.25$  & 20\%    & 0.9405485102 & 5 & 216.6 \\
	 $1$       & 8  & $\leq 0.20$ & 20\%    & 1.010395962 &  6 & 305.9\\
  	 $1$       & $8^*(3)$  & $\leq 0.70$ & 20\%    &  0.118671026 & 5  & 608.2\\
$1$       & 8  & $\leq 0.001$ & 100\%    & Infeasible  & 6 & 1.886  \\
	        \bottomrule
			\end{tabular}
	}

\end{tabular}
\vspace*{-0.2cm}
\caption{The scalability of our approach to pBNs with multiple parameters, detailed results for (left) \texttt{win95pts} and (right) \texttt{alarm}.} 
\label{tab:multi-par-results}
\vspace*{-0.4cm}
\end{table*}
\paragraph{RQ3: Sensitivity to the threshold $\lambda$.} 
We varied the constraint's threshold ($\lambda$) by steps of $\nicefrac{1}{20}$ for the benchmarks \texttt{alarm} \texttt{hepar2}, and \texttt{hailfinder} with $n{=}1$ and $\eta{=}99.99999\%$. 
Figures \ref{fig:threshhold-time}, \ref{fig:threshhold-distance}, and \ref{fig:threshhold-iteration} display the outcomes with the x-axis indicating the threshold and the y-axis indicating the tuning time (in seconds), the distance (log-scale), and the number of iterations.

\par \noindent \textbf{Findings:} By strengthening the threshold, the possibly satisfying regions get further away from $u_0$. Thus the distance, the number of iterations, and sometimes the tuning time grow. Similar findings are valid for $n>1$ parameters; see RQ4, Table \ref{tab:multi-par-results}. \textbf{Explanation:} Region refinement starts with small regions in the close vicinity of the original values of the parameters. Therefore, for the constraints close to the original probability $\Pr_{\B[u_0], \phi}$, the number of iterations is low, the distance is naturally small, and the minimal-change tuning is completed faster without the need to analyze larger regions.

\paragraph{RQ4: Scaling the number of parameters.} 
We took the \texttt{win95pts} and \texttt{alarm} benchmarks and parameterized them in multiple ways. 
Their pMCs have $9,947$ and $1,118$ states and $17,948$ and $3,147$ transitions respectively. 
The set of parameters for each pBN is including (and doubles the number of) parameters in the previous pBN.
Table~\ref{tab:multi-par-results}(left) and (right) list the results for \texttt{win95pts} and \texttt{alarm}. 
We list for each pBN, the number of affected CPTs, the number of parameters, the threshold $\lambda$, and the coverage $\eta$.
E.g., \texttt{win95pts} with $par{=}8$ has $8$ parameters occurring in 4 CPTs.
The columns EC, iter, and $t(s)$ report the EC-distance, the number of iterations, and the total time in seconds (incl.\ model building time, time for region refinement, and tuning time) respectively.
TO and MO indicate time-out (30 minutes) and memory-out ($>$16 GB).
 
 \par \noindent \textbf{Findings:} 
 (I) Approximately-close parameter tuning is feasible for pBNs with up to $8$ parameters. 
 \emph{This is significantly higher than the state of the art---one parameter.}
 As the number of sub-regions by PLA grows exponentially, treating more parameters is practically infeasible. 
 (II) More parameters often ease finding satisfying instantiations. E.g., the threshold $\geq 0.25$ is unsatisfiable for \texttt{win95pts} with $n{=}2$, but is satisfied with $n{=}4$. 
 (III) The results for multiple parameter pBNs confirm the findings for {RQ2} and {RQ3}; see the rows for each pBN instance with (i) varied coverage and (ii) varied threshold. 
 (IV) The unsatisfiability of a constraint can be computed fast (with $100\%$ confidence), regardless of the number of parameters, see e.g., \texttt{alarm} with $8$ parameters for the constraint $\leq 0.001$. 
 For infeasibility, a single verification suffices; no partitioning is needed. 


\vspace{0.2cm}
 \par \noindent  \textbf{RQ5: Handling pBNs with parameter dependencies.} Parameter lifting algorithm (Section {4.1}) enables handling models with parameter dependencies, see e.g., Example \ref{example-pBN-depdendency}: the parameters are allowed in multiple local distributions and the pBN sensitivity function is of higher degree. We extended our experiments to such cases for \texttt{win95pts} and \texttt{alarm}: we parameterized the entries $\theta_1 \cdots \theta_k \in \Theta_{\modif}$ over the same parameter $x \in X$ when the original values of $\theta_1 \cdots \theta_k$ were the same in the original BN. The eleventh row in Table \ref{tab:multi-par-results} (left) and the eighth and eleventh rows in Table \ref{tab:multi-par-results} (right) correspond to such cases. The term $8^*(2)(2)$ e.g., denotes that out of the $8$ parameters, 2 parameters repeatedly occurred in two distinct distributions. The term $8^*(3)$ denotes that out of the $8$ parameters, one was occurring in 3 distinct distributions.
 
 \par \noindent\textbf{Findings:} (I) Our method is applicable to pBNs with parameter dependencies where the sensitivity function is of a higher degree. (II) For the same coverage factor, the pBNs with parameter dependency are more expensive to analyze. See e.g., the two rows for $\texttt{win95pts}$ with $8$ parameters, threshold $\geq 0.115$, and the coverage $80\%$. This is due to more complex sensitivity functions that give a higher number of sub-regions to verify. (III) The pBNs with parameter dependency yielded notably smaller distances.

\vspace{-0.1cm}
\section{Epilogue}
\paragraph{Related work. }
Kwisthout and van der Gaag [\citeyear{DBLP:conf/uai/KwisthoutG08}] 
 studied the theoretical complexity of tuning problems. Renooij [\citeyear{DBLP:journals/ijar/Renooij14}] studied the properties of the co-variation schemes for BN tuning. She shows that the linear proportional scheme optimizes the CD distance for single-CPT pBNs. Similar are the studies by Bolt and van der Gaag [\citeyear{DBLP:conf/ecsqaru/BoltG15,DBLP:journals/ijar/BoltG17}] that consider tuning heuristics for distance optimizations. Peng and Ding [\citeyear{DBLP:conf/uai/PengD05}] propose an iterative proportional fitting procedure (IPFP) to minimize the KL-divergence distance for a set of constraints. The method does not scale to large networks. Santos \emph{et al.} [\citeyear{DBLP:journals/ijar/SantosGS13}] exploits linear programming for BN parameter tuning. Yak- aboski and Santos [\citeyear{DBLP:conf/webi/YakaboskiS18}] consider a new distance measure for parameter learning and tuning of BNs. Leonelli [\citeyear{DBLP:journals/ijar/Leonelli19}] considers nonlinear sensitivity functions, yet only for single parameter pBNs and Ballester-Ripoll and Leonelli [\citeyear{DBLP:conf/pgm/Ballester-Ripoll22}] efficiently compute the derivatives of sensitivity functions to select the most relevant parameters to a query, yet they limit to single parameter variation.

\paragraph{Conclusion.}A novel algorithm for parameter tuning in Bayesian networks is presented and experimentally evaluated.
Whereas existing algorithms come with severe restrictions---single parameters and/or linear functions---our approach is applicable to multiple (in practice about 8) parameters, large BNs (up to 100 variables), and polynomial functions. 
Future work includes considering balanced tuning heuristic \cite{DBLP:journals/ijar/BoltG17} and using monotonicity of parameters \cite{DBLP:conf/atva/SpelJK19}. 

\section*{Acknowledgement}
This research was funded by the ERC AdG Projekt
FRAPPANT (Grant Nr. 787914). We kindly thank Alexandra Ivanova for her implementation efforts and Tim Quatmann for the fruitful discussions.

\bibliographystyle{named}
\bibliography{ijcai23}

\end{document}